\documentclass[10pt,journal,compsoc]{IEEEtran}

\usepackage{amsmath,amsfonts}
\usepackage{algorithmic}
\usepackage{algorithm}
\usepackage{array}
\usepackage{textcomp}
\usepackage{stfloats}
\usepackage{url}
\usepackage{verbatim}
\usepackage{graphicx}
\usepackage{cite}
\usepackage{hyperref}
\usepackage{titletoc}
\usepackage{booktabs}
\usepackage{caption}
\usepackage{subcaption}
\usepackage{multirow}
\usepackage{amssymb}
\usepackage{bbding}
\usepackage{colortbl}
\usepackage[table]{xcolor}

\usepackage{orcidlink}

\definecolor{paired-light-blue}{RGB}{198, 219, 239}
\definecolor{paired-dark-blue}{RGB}{49, 130, 188}
\definecolor{paired-light-orange}{RGB}{251, 208, 162}
\definecolor{paired-dark-orange}{RGB}{230, 85, 12}
\definecolor{paired-light-green}{RGB}{199, 233, 193}
\definecolor{paired-dark-green}{RGB}{49, 163, 83}
\definecolor{paired-light-purple}{RGB}{218, 218, 235}
\definecolor{paired-dark-purple}{RGB}{117, 107, 176}
\definecolor{paired-light-gray}{RGB}{217, 217, 217}
\definecolor{paired-dark-gray}{RGB}{99, 99, 99}
\definecolor{paired-light-pink}{RGB}{222, 158, 214}
\definecolor{paired-dark-pink}{RGB}{123, 65, 115}
\definecolor{paired-light-red}{RGB}{231, 150, 156}
\definecolor{paired-dark-red}{RGB}{131, 60, 56}
\definecolor{paired-light-yellow}{RGB}{231, 204, 149}
\definecolor{paired-dark-yellow}{RGB}{141, 109, 49}  
\definecolor{myblue}{RGB}{218,232,252}
\definecolor{mygray}{RGB}{220,220,220}
\definecolor{mypink}{RGB}{251,49,153}

\newcommand{\myparagraph}[1]{\textbf{#1}\hspace{1.8ex}}

\newcommand{\jf}[1]{{\color{black} #1}}

\begin{document}

\title{\textit{MagicTime}: Time-lapse Video Generation Models as Metamorphic Simulators}

\author{
Shenghai~Yuan* ${\orcidlink{0009-0000-4850-2278}}$,
Jinfa~Huang* ${\orcidlink{0000-0002-0081-4106}}$,
Yujun~Shi ${\orcidlink{0009-0001-7594-7616}}$,
Yongqi~Xu ${\orcidlink{0009-0000-1635-6130}}$,
Ruijie~Zhu ${\orcidlink{0000-0003-4864-8474}}$,
\\
Bin~Lin ${\orcidlink{0009-0003-4805-9730}}$,
Xinhua~Cheng ${\orcidlink{0000-0001-9034-279X}}$,
Li~Yuan ${\orcidlink{0000-0002-2120-5588}}$,
Jiebo~Luo ${\orcidlink{0000-0002-4516-9729}}$, \textit{Fellow, IEEE}
\thanks{This work was supported in part by the Natural Science Foundation of China (No. 62202014, 62332002, 62425101, 62088102).}
\IEEEcompsocitemizethanks{
\IEEEcompsocthanksitem Shenghai Yuan, Yongqi Xu, Bin Lin, and Xinhua Cheng are with the Shenzhen Graduate School, Peking University, China. E-mail: \{yuanshenghai, xuyongqi, 2301212775, chengxinhua\}@stu.pku.edu.cn.
\IEEEcompsocthanksitem Yujun Shi is with the National University of Singapore, Singapore. E-mail: shi.yujun@u.nus.edu.
\IEEEcompsocthanksitem Ruijie Zhu is with the University of California, USA. E-mail: rzhu48@ucsc.edu.
\IEEEcompsocthanksitem Li Yuan is with the Shenzhen Graduate School, Peking University, China and Peng Cheng Laboratory, China. E-mail: yuanli-ece@pku.edu.cn.
\IEEEcompsocthanksitem Jiebo Luo and Jinfa Huang are with the Department of Computer Science, University of Rochester, USA. E-mail: \{jluo@cs, jhuang90@ur\}.rochester.edu.
\IEEEcompsocthanksitem  Corresponding authors: Li Yuan, Jiebo Luo; \ * means equal contributors.
}
\thanks{Manuscript received October 29, 2024; accepted March 27, 2025.}
\\
\bigskip
Code: \textcolor{red}{\url{https://pku-yuangroup.github.io/MagicTime}}
}

\markboth{IEEE TRANSACTIONS ON PATTERN ANALYSIS AND MACHINE INTELLIGENCE}%
{MagicTime: Time-lapse Video Generation Models as Metamorphic Simulators}


\IEEEtitleabstractindextext{%
\begin{abstract}
Recent advances in text-to-video generation (T2V) have achieved remarkable success in synthesizing high-quality general videos from textual descriptions. A largely overlooked problem in T2V is that existing models have not adequately encoded physical knowledge of the real world, thus generated videos tend to have limited motion and poor variations. In this paper, we propose \textit{MagicTime}, a metamorphic time-lapse video generation model, which learns real-world physics knowledge from time-lapse videos and implements metamorphic generation. First, we design a simple yet effective two-stage Magic Adaptive Strategy, encode more physical knowledge from metamorphic videos, and transform pre-trained T2V models to generate metamorphic videos. Second, we introduce a Dynamic Frames Extraction strategy to adapt to metamorphic time-lapse videos, which have a wider variation range and cover dramatic object metamorphic processes, thus embodying more physical knowledge than general videos. Finally, we introduce a Magic Text-Encoder to improve the understanding of metamorphic video prompts. Furthermore, we create a time-lapse video-text dataset called \textit{ChronoMagic}, specifically curated to unlock the metamorphic video generation ability. Extensive experiments demonstrate the superiority and effectiveness of MagicTime for generating high-quality and dynamic metamorphic videos, suggesting time-lapse video generation is a promising path toward building metamorphic simulators of the physical world.
\end{abstract}

}

\maketitle

\begin{figure*}[!t]
    \centering
    \includegraphics[width=0.9\linewidth]{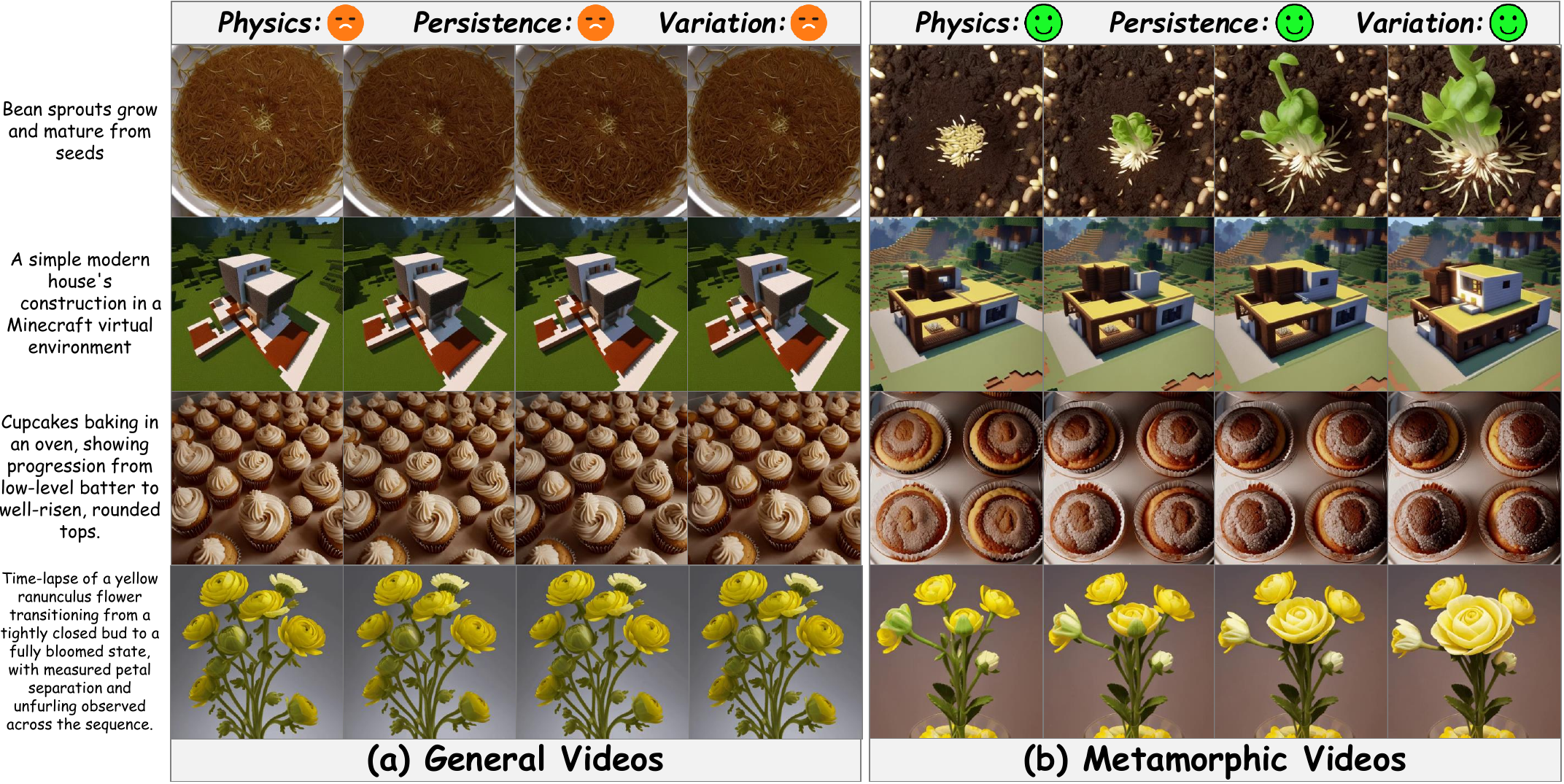}
    \caption{\textbf{The illustration of the difference between (a) general videos, and (b) metamorphic videos.} Compared to general videos, metamorphic videos contain physical knowledge, long persistence, and strong variation, making them difficult to generate. The general videos are generated by the Animatediff \cite{animatediff}. The metamorphic videos generated by our MagicTime demonstrate that it can greatly improve the capabilities of metamorphic video generation.
    }
    \label{figure_def}
\end{figure*}

\section{Introduction}\label{sec: intro}
\IEEEPARstart{R}{ecent} advances in text-to-video (T2V) generation models have been driven by the emergence of diffusion models. These models are typically based on either Transformer architectures \cite{Videopoet, Latte} or U-Net architectures \cite{VideoComposer, Modelscope, Align_your_latents, animatediff}. The former approach tokenizes videos and incorporates positional encoding for training, whereas the latter extends the 2D U-Net with a temporal feature extraction module for pseudo-3D training. However, the videos generated by these models primarily exhibit camera movement, lacking a continuous subject metamorphic process and limited incorporation of physical knowledge. We categorize such videos as \textit{general videos}, illustrated in Fig. \ref{figure_def}a. Since the training datasets predominantly consist of general videos, current T2V models struggle to produce videos depicting complex phenomena like seed germination, ice melting, or flower blooming, which are characterized by minimal physical content, brief duration, and limited variation.
\begin{table}[t]
\centering
  \setlength\tabcolsep{0.85mm}
\caption{\textbf{Comparison of the characteristics of our ChronoMagic benchmark with existing time-lapse datasets.} We compare time-lapse video characteristics in terms of physics, persistence, and variation.}
\begin{tabular}{c|ccc}
\toprule
\textbf{Dataset} & \textbf{Strong Physics} & \textbf{High Persistence} & \textbf{Variation}\\
\midrule
Sky Time-Lapse \cite{SkyTimelapse}  & \XSolidBrush & \XSolidBrush & \XSolidBrush \\
Time-Lapse-D \cite{TimeLapse-D}  & \XSolidBrush & \XSolidBrush & \XSolidBrush \\
\rowcolor{myblue} ChronoMagic  & \CheckmarkBold & \CheckmarkBold & \CheckmarkBold \\
\bottomrule
\end{tabular}
\label{tab:dataset_com}
\end{table} 

Compared to general videos, we have observed a category of videos that typically encompasses the subject's entire transformation process, thus addressing the inherent limitations of the former. We term this category as \textit{metamorphic videos}, which encode a more comprehensive representation of world knowledge, as illustrated in Fig. \ref{figure_def}b. Such videos have the potential to enhance the capacity of video generation models to accurately depict the real world. Consequently, the generation of metamorphic videos is of paramount importance for a wide range of applications, despite being relatively underexplored in existing research. While the study \cite{freebloom} demonstrates the ability to create similar effects by repeatedly inferring and concatenating general videos, it falls short in encoding physical knowledge and lacks generalization through a prompt strategy. In contrast, we aim to develop an end-to-end time-lapse video generation method for adaptively encoding physical knowledge with strong generalization capabilities.

Time-lapse videos provide detailed documentation of an object's complete metamorphosis, possessing the essential characteristics of metamorphic videos. Consequently, we utilize time-lapse videos to validate our hypothesis. A straightforward approach is to leverage existing powerful T2V models to directly generate metamorphic videos. Since metamorphic videos and general videos have intrinsic gaps, current methods struggle to produce satisfactory metamorphic videos directly. The core challenges include: \textbf{(1)} Metamorphic time-lapse videos contain more physics, requiring stronger structural extraction; \textbf{(2)} It has longer persistence, fully covering object metamorphosis and being sensitive to training strategies; \textbf{(3)} Metamorphic time-lapse videos have greater variation, potentially transforming from seed to flower, demanding high-quality data. 

To tackle the challenge of generating metamorphic videos, we introduce a novel framework, \textbf{MagicTime}, which excels in creating video content that is \textbf{Magic}ally compressed in \textbf{Time}. Initially, we present the Magic Adaptive Strategy to encode more physical knowledge in metamorphic videos into the feature extraction, expanding the metamorphic video generation capabilities of the pre-trained T2V model. Subsequently, we propose a Dynamic Frames Extraction strategy that enables the model to adapt to the characteristics of time-lapse training videos and prioritize metamorphic video features without altering its core general video generation abilities. Moreover, we introduce a Meta Text-Encoder to refine prompt understanding and distinguish between metamorphic and general prompts. As shown in Fig. \ref{figure_def}b, our MagicTime generates text-aligned, coherent, high-quality metamorphic videos across various styles, coordinated with carefully designed techniques. 

Further, we meticulously collect a dataset, \textbf{ChronoMagic}, sourced from Internet time-lapse videos for training our method. As shown in Table \ref{tab:dataset_com}, our ChronoMagic dataset is distinct from previous time-lapse datasets \cite{SkyTimelapse, TimeLapse-D}, as it consists of metamorphic time-lapse videos (e.g. ice melting, and flower blooming) that show strong physics, high persistence, and variation. Given the domain discrepancy between metamorphic and general videos, we propose the integration of Cascade Preprocessing and Multi-View Text Fusion to capture dynamics into detailed video captions. Contributions are summarized as follows:
\begin{itemize}
    \item We introduce \textbf{MagicTime}, a metamorphic time-lapse video generation diffusion model, by adopting a standard T2V model with a Magic Adaptive Strategy. This addresses the limitations of existing methods that are unable to generate metamorphic videos.
    \item We design a Dynamic Frames Extraction strategy to effectively extract physics from metamorphic videos and a Magic Text-Encoder to enhance the model's understanding of metamorphic text prompts.
    \item We propose an automatic metamorphic video captioning annotation framework. Utilizing this framework, we have curated a high-quality dataset named \textbf{ChronoMagic}, consisting of 2,265 time-lapse videos, each accompanied by a detailed caption.
    \item Extensive experiments demonstrate that our \textbf{MagicTime} is capable of generating high-quality and consistent metamorphic time-lapse videos.
\end{itemize}

\section{Related Work}
\myparagraph{Text-to-Image Generation}
Text-to-image (T2I) synthesis has made significant strides thanks to advancements in deep learning and the availability of large image-text datasets such as CC3M \cite{CC3M} and LAION \cite{LAION}. Compared to variational autoencoders (VAEs) \cite{VAE, vqVAE}, flow-based models \cite{Rezende_Mohamed_2015, Dinh_Sohl-Dickstein_Bengio_2016}, and generative adversarial networks (GANs) \cite{tedigan, StyleGAN-XL, StyleGAN-T}, diffusion models \cite{Imagen, DALLE, DALLE2, sdxl, lian2023llm, Multidiffusion} have demonstrated exceptional efficacy in image modeling. For instance, contemporary T2I models often employ Latent Diffusion Models (LDM) \cite{LDM}, which utilize sampling in latent spaces from pre-trained autoencoders to minimize computational overhead. Recapti \cite{Recapti} harnessed the power of Large Language Models (LLM) to improve the text comprehension capabilities of LDM, enabling training-free T2I generation. Furthermore, the integration of additional control signals \cite{ucontrolnet, chen2024training, t2iadapter, dragdiffusion, Gligen} has further expanded the applicability of Diffusion Models in T2I generation. Unlike image generation, the task of video generation required by MagicTime is more complex, as it necessitates both diversity in content across space and consistency in content across time.

\begin{figure*}[!t]
  \centering
  \includegraphics[width=0.9\linewidth]{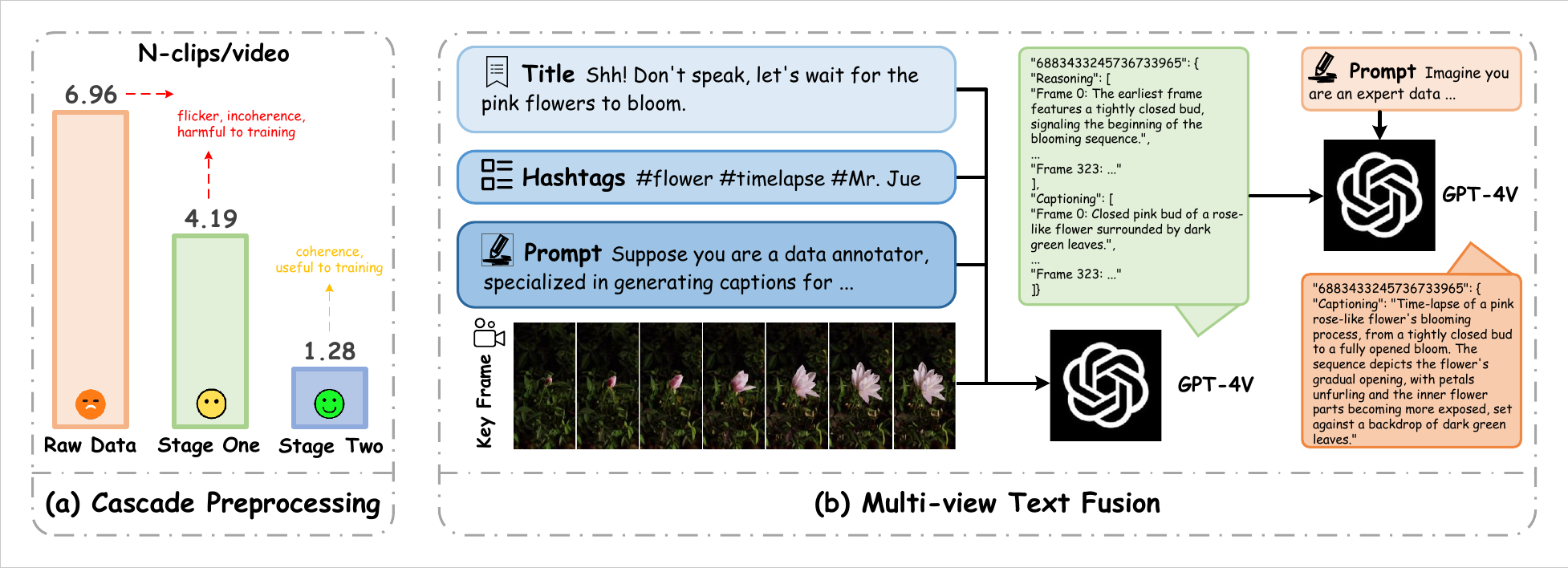}
  \caption{\textbf{The pipeline for high-quality time-lapse video data annotation. (a) Cascade Preprocessing.} \jf{
 It automatically divides a video into clips at corresponding transitions in each stage, thus dividing a complex raw video into clips. N-clips/video refers that a video needs to be divided into N clips, and the smaller the value, the more coherent the video.}     \textbf{(b) Multi-view Text Fusion.} Compared with only inputting keyframes, our pipeline can obtain vivid, detailed, and accurate descriptions.
  }
  \label{figure_multiview}
\end{figure*}

\noindent \myparagraph{Text-to-Video Generation}
The advent of large-scale video-text datasets \cite{webvid, Internvid, HD-VG-130M} has facilitated significant advancements in pre-trained video diffusion models \cite{zhang2024motiondiffuse, Make-a-video, SVD, Align_your_latents, ge2023preserve, Modelscope, Show-1, Followyourpose, Follow-your-click, Follow-your-emoji}. MagicVideo2 \cite{MagicVideo-V2} utilized cascaded diffusion to separate image and video generation, starting with text to initially generate images and subsequently produce high-quality videos. Videopoet \cite{Videopoet} employed a Transformer-based diffusion approach, combined with multimodal inputs, for zero-shot video generation. Animatediff \cite{animatediff}, and Lumiere \cite{lumiere} extended the 2D-UNet to a 3D-UNet by adding temporal layers and implementing a latent denoising process to conserve computational resources. While these models have achieved basic video generation, the resulting videos typically feature a single scene and a single action, and lack encoding of physical world knowledge. In contrast, our MagicTime incorporates more physics knowledge and offers the capability to generate metamorphic videos.

\noindent \myparagraph{Time-lapse Video Generation}
Previous studies~\cite{nam2019end,FGLA, DtvNet, AL} on time-lapse video generation predominantly concentrated on general videos, failing to produce the metamorphic videos that we address. For example, \cite{FGLA, DtvNet} treated time-lapse videos merely as a distinct domain, overlooking the physical knowledge embedded within them. Consequently, the generated videos were limited to scenes with relatively minor movements, such as drifting clouds or moving sunsets. In contrast, we primarily focus on generating metamorphic time-lapse videos, which exhibit significant variations and depict the complete process of object metamorphosis. The goal of our work is to integrate more physical knowledge into MagicTime for the generation of high-quality metamorphic time-lapse videos.

\section{Methodology}
\begin{figure*}[!t]
  \centering
  \includegraphics[width=0.9\linewidth]{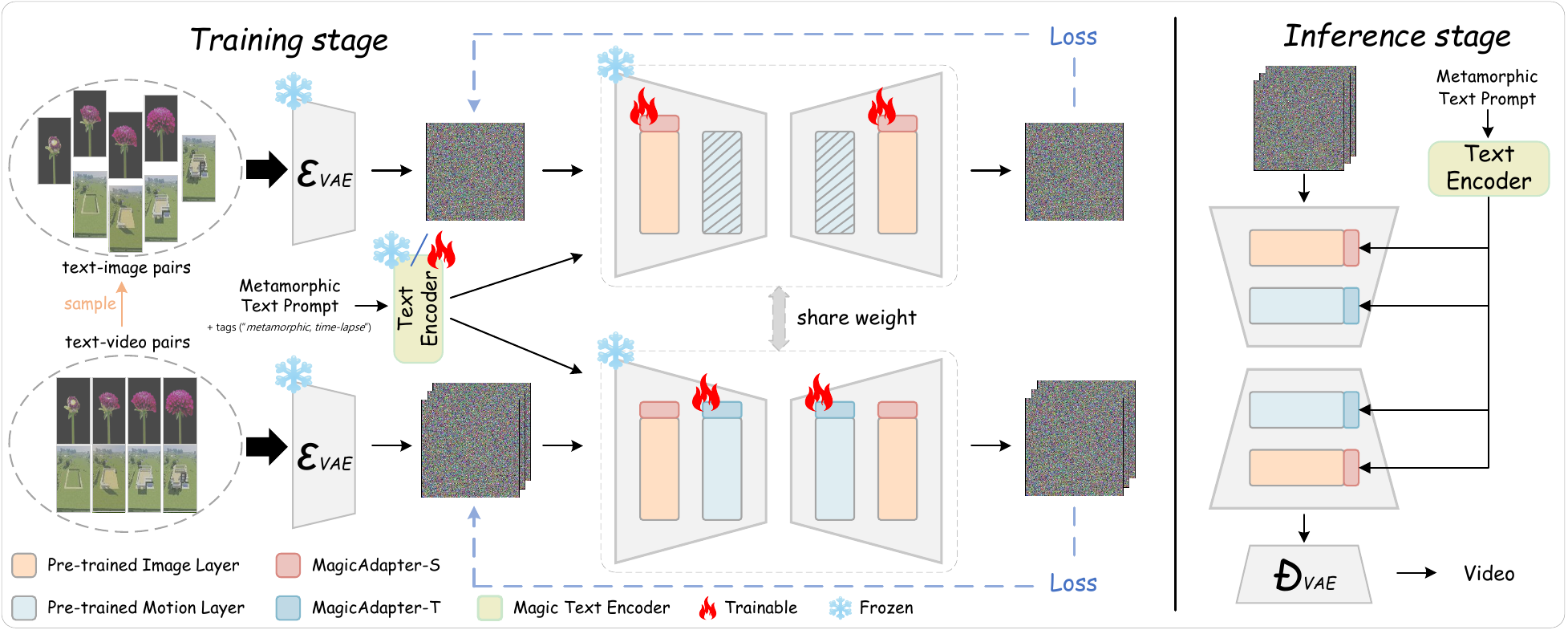}
  \caption{\textbf{Overview of the proposed MagicTime approach.} We first train  \textit{MagicAdapter-S} to focus on static state information. Next,  \textit{MagicAdapter-T} is trained to encode dynamic stage change with the help of \textit{Dynamic Frames Extraction}. Finally, we train a \textit{Magic Text-Encoder} to enhance text comprehension ability. During the inference stage, all components need to be loaded simultaneously. Slash padding indicates the module is not used.}
  \label{figure_pipeline}
\end{figure*}

This section begins with a brief overview of the diffusion model, followed by a description of the ChronoMagic dataset construction. Finally, we provide an outline of MagicTime (see Fig. \ref{figure_pipeline}) and discuss its implementation details.

\subsection{Preliminaries}
Diffusion models \cite{DDPM, DDIM, grosz2024universal} typically consist of two processes: forward diffusion and reverse denoising. Given an input signal $x_t$, the process of forward diffusion is defined: 
\begin{equation}
p_\theta\left(x_t \mid x_{t-1}\right)=\mathcal{N}\left(x_t ; \sqrt{1-\beta_{t-1}} x_{t-1}, \beta_t I\right),
\end{equation} 
where $t = 1, ..., T$ are time steps, and $\beta_t \in (0, 1)$ is the noise schedule. When the total time steps $T$ are sufficiently large, the resulting $x_t$ eventually approximates Gaussian noise $\in N(0, I)$. The purpose of the denoising process is to learn how to reverse the forward diffusion:
\begin{equation}
p_\theta\left(x_{t-1} \mid x_t\right)=\mathcal{N}\left(x_{t-1} ; \mu_\theta\left(x_t, t\right), \Sigma_\theta\left(x_t, t\right)\right),
\end{equation}
where $\Sigma_\theta(x_t, t)$ often does not need to be trained but is calculated based on the time step $t$ as a fixed variance. It only needs to predict the mean $\mu_\theta(x_t, t)$ in the reverse process, and this step can also be simplified to training the denoising model $\epsilon_\theta(x_t, t)$ to predict the noise $\epsilon$ of $x_t$:
\begin{equation}
\mathcal{L}=\mathbb{E}_{x_t, y, \epsilon \sim \mathcal{N}(0, I), t}\left[\left\|\epsilon-\epsilon_\theta\left(x_t, t, \tau_\theta(y)\right)\right\|_2^2\right],
\end{equation}
where $y$ is text condition, and $\tau_\theta(\cdot)$ is the text encoder. By replacing $x_t$ with $\mathcal{E}\left(x_t\right)$, the latent diffusion model is derived \cite{LDM, li2024latent}, which also serves as the loss function for MagicTime. In video diffusion models, the parameter $\theta$ typically employs either a Transformer \cite{Videopoet, Latte} or a U-Net architecture \cite{shen2024holi, xia2024diffusion, luo2024measurement, sun2024create, VideoComposer, Modelscope, Align_your_latents, animatediff}. Considering that the most effective open-source T2V models currently utilize U-Net, we have embraced this architecture in this work.

\subsection{ChronoMagic Dataset}
\myparagraph{Data Curation and Filter}\label{sec:data-curation-filter} We found that videos in existing large-scale text-video datasets like WebVid~\cite{webvid} typically consist of coherent segments depicting everyday life (general videos), devoid of significant changes or object metamorphosis. Consequently, open-source models \cite{SVD, Make-a-video, animatediff} are limited to generating general videos and are unable to produce metamorphic videos. To overcome this limitation, we introduce the ChronoMagic dataset, which includes 2,265 high-quality time-lapse videos with accompanying text descriptions from open-domain content. Time-lapse videos capture complete object metamorphosis and convey more physical knowledge than general videos, thereby enhancing generative models' understanding of the world.

Given that video platforms host a multitude of low-quality videos, indiscriminate collection would inevitably compromise the dataset's quality. To mitigate this, we leverage the metadata from video websites for collection and filtering. Our data collection process begins with retrieving original videos from YouTube using "time-lapse" as a search term. Subsequently, we implement a filtering mechanism that excludes videos with short titles, low view counts, or absent hashtags. Furthermore, we discard videos with irrelevant hashtags such as "YouTube," "video," and "shorts." Building on the annotations generated in Section~\ref{sec:multi-view-text-fusion}, we further refine our selection by eliminating irrelevant videos, thus establishing a closed-loop data collection and iteration process. Ultimately, we curate a collection of 2,265 time-lapse videos that meet our specified criteria.

\noindent \myparagraph{Cascade Preprocessing}\label{sec:CasacdePreprocess} Due to complex scene transitions in Internet videos (see Fig. \ref{figure_multiview}a) affecting the model's learning of temporal relevance, we need to find a way to solve this problem. We first define $F$ as the set of all video frames. In the initial stage, each frame $f_{i}$ in $F$ is converted to grayscale $g_{i} \in (1, H, W)$ to mitigate the influence of color on detection. Subsequently, we calculate the pixel intensity difference $\mathcal{D}_{i, i+1}$ between consecutive frames, yielding the average pixel intensity $\mathcal{S}_{i, i+1}$:
\begin{equation}\mathcal{D}_{i, i+1}=\left|g_i-g_{i+1}\right|,\quad\mathrm{for~}i=1\mathrm{~to~}F-1
\end{equation}
\begin{equation}\mathcal{S}_{i, i+1}=\sum_{h=1}^H\sum_{w=1}^W\mathcal{D}_{i, i+1}(h,w),\quad\mathrm{for~}i=1\mathrm{~to~}F-1
\end{equation}
if $\mathcal{S}_{i,i+1} > \theta$, then this position is defined as the transition point. However, we noted that this simple method is prone to occasional errors. Consequently, in stage two, we further employ CLIP~\cite{CLIP} to detect transitions:
\begin{equation}\mathcal{S}_{i,i+1}^{clip}=\text{cos}(\operatorname{CLIP}(f_i),\operatorname{CLIP}(f_{i+1})),\quad\mathrm{for~}i=1\mathrm{~to~}F-1
\end{equation}
if $\mathcal{S}_{i,i+1}^{clip} < \vartheta$, then this position is defined as the transition point. The $\text{cos} (\cdot)$ returns a similarity vector. After that, we use a voting mechanism to decide whether a specific location should be recognized as a transition point:
\begin{equation}
\mathcal{T}_i = \begin{cases} 
\text{True,} & \text{if } \mathcal{S}_{i, i+1} > \theta \text{ and } \mathcal{S}_{i,i+1}^{clip} < \vartheta \\
\text{False,} & \text{otherwise}
\end{cases}
\end{equation}
where $i$ represents the position of the frame. Ultimately, we segment the video into distinct sections using the identified transition points. As illustrated in Fig. \ref{figure_multiview}a, this process results in a high-quality video dataset with fewer transitions.

\noindent \myparagraph{Multi-view Text Fusion}\label{sec:multi-view-text-fusion} 
After preprocessing, the videos must be annotated with appropriate captions. A simple method involves feeding the video into a multimodal model \cite{Video-llava, PLLaVA, LLaVA-NeXT, chen2024dynamic} for video description. However, we observed a significant occurrence of hallucination in open-source models, as described in the Appendix. To address this issue, we adopted In-Context Learning and Chain-of-Thought \cite{ICL} based on GPT-4V \cite{gpt4, lyu2023gpt} to generate captions in stages. Fig. \ref{figure_multiview}b shows the detailed text generation and fusion process. During this process, we also used titles and hashtags as supplements to reduce model biases. Specifically, we initially employ GPT-4V \cite{gpt4} to generate keyframe captions with the help of titles and hashtags. Subsequently, these captions are used to develop a comprehensive representation of the entire video, culminating in the generation of the final video captions. Additionally, at this stage, GPT-4V evaluates whether the video is a time-lapse and then forms a closed data loop, as described in Section ~\ref{sec:data-curation-filter}.

\begin{figure}[t]
  \centering
  \includegraphics[width=1\linewidth]{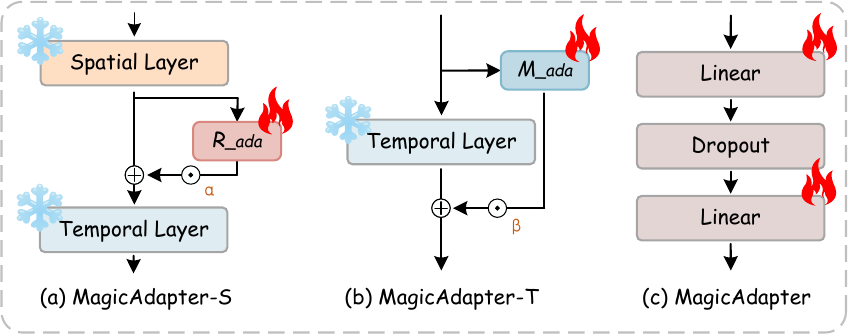}
  \caption{\textbf{The structure of MagicAdapter and its functional position.} MagicAdapter consists of several layers of Linear layers. MagicAdapter-S operates on the output of the Spatial Layer, while MagicAdapter-T functions on the input of the Temporal Layer. $\alpha$ and $\beta$ represent the weight coefficients.}
  \label{figure_magicadapter}
\end{figure}

\subsection{MagicTime Training Recipe}\label{method}

\myparagraph{Magic Adaptive Strategy}The key to encoding metamorphic physical priors into the model lies in decoupling the training process. Video generation models typically consist of spatial layers and temporal layers. The former focuses on learning visually relevant content, while the latter aims to establish motion patterns. To enable the pre-trained text-to-video (T2V) model to generate metamorphic videos, we divide the training process into two phases: spatial training and temporal training, as shown in Fig. \ref{figure_pipeline}.

The reasons and benefits are as follows: First, compared to general videos, metamorphic videos contain more static state information (e.g., seeds, leaves, flowers) and dynamic state change information (e.g., seed $\rightarrow$ leaf $\rightarrow$ flower). The decoupled training process can separately encode these two characteristics, thus better encoding new physical priors without losing the original priors. Secondly, the phased training strategy improves the robustness of the model and exhibits better physical consistency when generating videos. Additionally, it reduces training complexity, allowing the model to focus on spatial details and temporal changes separately, thereby improving training stability and efficiency.

To achieve this goal, in the first phase, we remove the temporal layer from the pre-trained model and integrate \textit{MagicAdapter-S} into the spatial layer. Subsequently, we freeze the remaining model parameters and train using keyframe-text pairs from the ChronoMagic dataset. Since this phase does not involve temporal changes, the model can learn these static features more accurately and also alleviate the impact of video watermarks. In the second phase, we reintroduce the pre-trained temporal layer and propose integrating \textit{MagicAdapter-T} into the temporal layer. We then freeze other parameters and train using video-text pairs from the ChronoMagic dataset. Since spatial features have been learned in the previous phase, the optimization during the temporal training phase can focus more on the rules of metamorphic changes, thereby avoiding mutual interference between the learning of spatial and temporal features. Our proposed \textit{MagicAdapter} is simple and lightweight, as shown in Fig. \ref{figure_magicadapter}. Formally, let \(X \in \mathbb{R}^{ \text{B} \times \text{C} \times \text{1} \times \text{H} \times \text{W}} \) and \( Y \in \mathbb{R}^{ \text{B} \times \text{C} \times \text{F} \times \text{H} \times \text{W}} \) be the input to a single spatial layer \( \mathcal{F}_{sap} \) and \textit{MagicAdapter-S} layer \( \mathcal{R}_{ada} \), and to a single temporal layer \( \mathcal{F}_{temp} \) and \textit{MagicAdapter-T} layer \( \mathcal{M}_{ada} \).
\begin{equation}
\quad \mathcal{F}_{outs}(X) = \mathcal{F}_{sap}(X) + \alpha \times \mathcal{R}_{ada}(\mathcal{F}_{sap}(X)),
\end{equation}
\begin{equation}
\quad \mathcal{F}_{outt}(Y) = \mathcal{F}_{temp}(Y) + \beta \times \mathcal{M}_{ada}(Y),
\end{equation}
where \( \mathcal{F}_{outs} \in \mathbb{R}^{\text{B} \times \text{C} \times \text{1} \times \text{H} \times \text{W}} \) and \( \mathcal{F}_{outt} \in \mathbb{R}^{\text{B} \times \text{C} \times \text{F} \times \text{H} \times \text{W}} \), \(\alpha\) and \(\beta\) are scaling coefficients, \( B \), \( C \), \( F \), \( H \), and \( W \) represent the batch size, channel, frame number, height, and width.

\noindent \myparagraph{Dynamic Frames Extraction}\label{sec:DFE} 
Open-source algorithms, whether for video generation \cite{360DVD, Magic-Me} or multimodal models \cite{Video-llava, Languagebind}, typically sample a random continuous sequence of $N$ frames for training. However, this approach captures only a brief segment of the full video, leading to limitations such as monotonous actions and restricted variation in amplitude. This is not conducive to extracting training frames for metamorphic videos. By analyzing ChronoMagic's characteristics, we opt to sample $N$ frames evenly from a video, capitalizing on the nature of metamorphic videos that encompass the entire process of object metamorphosis. This ensures that the training data exhibit metamorphic properties, rather than being confined to general ones, thereby enhancing the generative models' ability to comprehend physical phenomena. Nonetheless, experiments indicate that uniform sampling from all videos may compromise the model's original capabilities, resulting in severe flickering phenomena in the generated videos. Therefore, we retain a portion of general videos and dynamically select the sampling strategy based on the results of \textit{cascade preprocessing} to avoid catastrophic forgetting:
\begin{equation}
    \mathcal{S}^{*} = \begin{cases} 
    \text{Uniform, with } \mathcal{P} \text{, if } \sum_{i=1}^{F-1}\mathcal{T}_i \leq \delta \\
    \text{Random, with } \mathcal{P} \text{, if } \sum_{i=1}^{F-1}\mathcal{T}_i > \delta
    \end{cases}
\end{equation}
where probability $\mathcal{P}$ represents the likelihood of executing the corresponding sampling strategy, enhancing model robustness. $\delta$ represents the threshold for judging transition.

\begin{table}[t]
    \centering
    \caption{\textbf{Comparison of statistics among \textit{Sky Time-lapse} \cite{SkyTimelapse}, \textit{Time-lapse-D} \cite{TimeLapse-D}, and the proposed metamorphic \textit{ChronoMagic} dataset.} \textit{ChronoMagic} is the only open-domain metamorphic time-lapse dataset with video-text pairs.}
    \label{table_dataset_2}
    \resizebox{\columnwidth}{!}
    {
    \begin{tabular}{c|ccc}
        \toprule
                     \textbf{Dataset}                  & \textbf{ChronoMagic} & \textbf{Sky Time-lapse} & \textbf{Time-lapse-D} \\
        \midrule
        \textbf{Type}                  & Video-Text & Video & Video \\
        \textbf{Category}              & Open-domain & Sky & Sky \\
        \textbf{Variation}             & Strong (metamorphic) & Weak (general) & Weak (general) \\
        \textbf{Clips}                 & 2,265 & 2,049 & 16,874 \\
        \textbf{Resolution}            & 1920 $\times$ 1080  & 640 $\times$ 360 & 1920 $\times$ 1080  \\
        \textbf{Information Density}   & Dense & Sparse & Sparse \\
        \bottomrule
    \end{tabular}
    }
\end{table}

\noindent \myparagraph{Magic Text-Encoder}
Textual descriptions corresponding to metamorphic videos typically contain more temporal and state information than those for general videos. Consequently, using a text encoder trained exclusively on general videos is deemed insufficient for this purpose. To overcome this limitation, we introduce the Magic Text-Encoder, which is specifically designed to encode metamorphic prompts while retaining its ability to process general prompts. We select CLIP \cite{CLIP} as our initial text encoder and incorporate several convolution layers to encode metamorphic information while ensuring high efficiency, inspired by Lora \cite{Qlora}. During training, we incorporate the tags "metamorphic, time-lapse" into the text prompts corresponding to the metamorphic videos. This approach emphasizes the difference between metamorphic and general videos, allowing Magic Text-Encoder to incorporate metamorphic priors without losing its original text-following ability. The training is conducted after the overall network training is completed, because this step is to further refine the model to better interpret the nuances of the metamorphic. Formally, given an original projection \(\mathbf{X}_0 \mathbf{W}_0 = \mathbf{Y}_0\) where \(\mathbf{X}_0 \in \mathbb{R}^{q \times w}\) and \(\mathbf{W}_0 \in \mathbb{R}^{w \times r}\), the Magic Text-Encoder computes:
\begin{equation}
\mathbf{Y}_1 = (\mathbf{X}_0 + \mathbf{T}_{\text{AG}}) \mathbf{W}_0 + \gamma (\mathbf{X}_0 + \mathbf{T}_{\text{AG}}) \mathbf{L}_1 \mathbf{L}_2,
\end{equation}
where \(\mathbf{X}_0\) is the input text prompt of CLIP, \(\mathbf{Y}_1\) is the output of CLIP, \(\mathbf{T}_{\text{AG}}\) indicates newly added metamorphic tags, \(\mathbf{L}_1 \in \mathbb{R}^{w \times e}\) and \(\mathbf{L}_2 \in \mathbb{R}^{e \times r}\) are adapter parameters, \(\mathbf{W}_0\) are the CLIP parameters, \(\gamma\) is a scaling coefficient, \(q\), \(w\), \(e\), and \(r\) represent the dimensions of input, intermediate, low-rank, and output, respectively. To avoid losing original capabilities, only \(\mathbf{L}_1\) and \(\mathbf{L}_2\) are updated. More details of the structure of Magic Text-Encoder can be found in the Appendix.



\section{Experiment}\label{sec:experiment}

\begin{figure*}[!t]
    \centering
    \includegraphics[width=0.9375\linewidth]{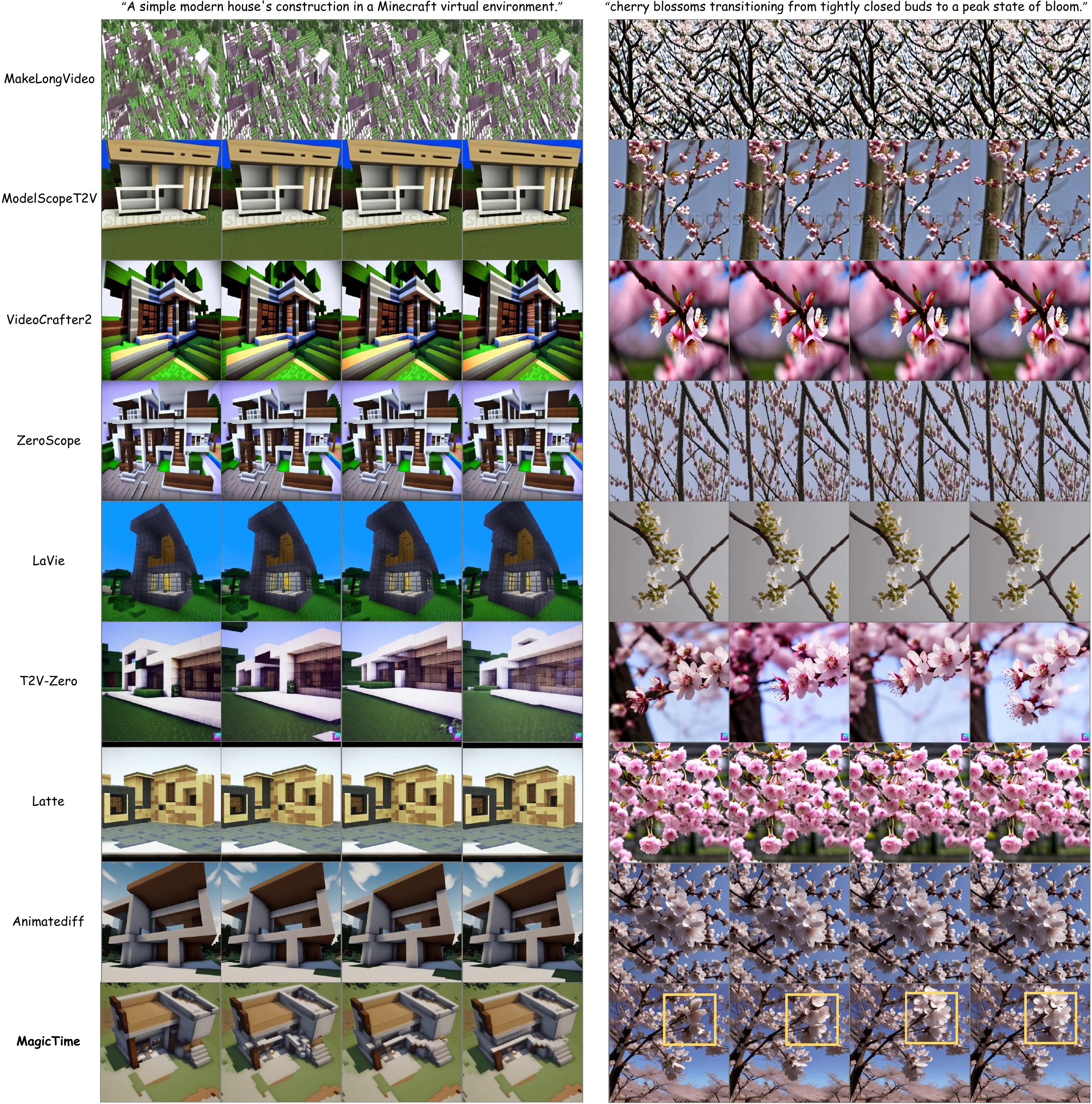}
    \caption{\textbf{Qualitative comparison for metamorphic video generation}. Among \textit{MakeLongVideo} \cite{MakeLongVideo}, \textit{ModelScopeT2V} \cite{Modelscope}, \textit{VideoCrafter2} \cite{VideoCrafter2}, \textit{Zeroscope} \cite{zeroscope}, \textit{LaVie} \cite{lavie}, \textit{T2V-Zero} \cite{T2V-zero}, \textit{Latte} \cite{Latte} and \textit{Animatediff} \cite{animatediff}, MagicTime achieves best results.
    }
    \label{figure_comparision_CTL_1}
\end{figure*}

\subsection{Experimental Settings} \label{sec:exp setting}

\myparagraph{Datasets}
Metamorphic video generation represents a novel paradigm, yet dedicated evaluation datasets are scarce, presenting challenges for thorough method assessment. To address this need, we utilize the ChronoMagic dataset, the first benchmark specifically designed for metamorphic video generation. Detailed information and examples are provided in the Appendix. As shown in Table \ref{table_dataset_2}, the ChronoMagic dataset is the only \textit{metamorphic time-lapse} dataset with \textit{video-text} pairs, which has more categories and stronger variation, that is, higher physical content. Despite the seemingly lower number of videos for ChronoMagic, our overall information density is higher than that of \textit{Sky Time-lapse} \cite{SkyTimelapse} and \textit{Time-lapse-D} \cite{TimeLapse-D}, as they mainly consist of general time-lapse videos such as cloud drifting.




\noindent \myparagraph{Metrics}
In this work, we utilize three metrics for quantitative evaluation: Fréchet Inception Distance (FID) \cite{FID}, Fréchet Video Distance (FVD) \cite{FVD} and CLIP Similarity (CLIPSIM) \cite{CLIPSIM}. These are standard indicators assessing generated image quality, video quality, and text-content similarity. For fairness, we fix the random seed in all experiments.

\noindent \myparagraph{Implementation Details}
We choose Stable Diffusion v1.5 \cite{LDM} and Motion Module v1.5 \cite{animatediff} as our baseline models. In the training stage, the resolution is set to 512$\times$512, 16 frames are taken by Dynamic Frames Extraction from each video, the batch size is set to 1,024, the learning rate is set to 1e-4, the total number of training steps is 10k, and the classify free guidance random null text ratio is 0.1. We use a linear beta plan as the optimizer with $\beta_{\text{start}}=8.5e-4$ and $\beta_{\text{end}}=1.2e-2$. We set $\theta$, $\vartheta$, $\mathcal{P}$ and $\delta$ in Sections \ref{sec:CasacdePreprocess} and \ref{sec:DFE} to 40, 0.5, 0.9 and 3 respectively to enhance robustness. In the inference stage, we utilize DDIM with 25 sampling steps, a text-guided scale of 8.0, and a sampling size of 512$\times$512. We have collected personalized SD models from CivitAI to verify the generalizability of our method, including RealisticVision, RcnzCartoon, and ToonYou. Due to space limitations, more qualitative experiments on generated metamorphic videos are shown in Appendix. Further, selected examples from our ChronoMagic dataset, along with the questionnaire employed for human evaluation, are also provided in the Appendix for further reference.

\begin{table}[!t]
    \centering
    \caption{\textbf{Quantitative comparison with state-of-the-art T2V generation methods for the text-to-video task.} Results are produced on ChronoMagic. Our method achieves either the best results or second-best results across all metrics. "$\downarrow$" denotes lower is better. "$\uparrow$" denotes higher is better.}
    \resizebox{\columnwidth}{!}
    {
    \begin{tabular}{lccccc}
    \toprule
    \textbf{Method} & \textbf{Venue} & \textbf{Backbone}  & \textbf{FID$\downarrow$} & \textbf{FVD$\downarrow$} & \textbf{CLIPSIM$\uparrow$} \\
    \midrule
    MakeLongVideo \cite{MakeLongVideo} & Github'23 & U-Net & 143.07 & 511.62 & 0.2852 \\
    ModelScopeT2V \cite{Modelscope} & Arxiv'23 & U-Net & 61.49 & 500.43 & 0.3093 \\
    VideoCrafter2 \cite{VideoCrafter2} & Arxiv'24 & U-Net & \textbf{54.61} & 473.69 & 0.3149 \\
    Zeroscope \cite{zeroscope} & CVPR'23 & U-Net & 103.74 & 497.60 & 0.3079 \\
    LaVie \cite{lavie} & Arxiv'23 & U-Net & 62.28 & \underline{445.60} & 0.3090 \\
    T2V-zero \cite{T2V-zero} & ICCV'23 & U-Net & 58.58 & 488.85 & \underline{0.3203} \\
    Latte \cite{Latte} & Arxiv'24 & Transformer & 59.10 & 445.85 & 0.2939 \\
    Animatediff \cite{animatediff} & ICLR'24 & U-Net & 63.83 & 451.35 & 0.3156 \\
    \rowcolor{myblue} \textbf{MagicTime} & \textbf{Ours} & U-Net & \underline{58.12} & \textbf{441.17} & \textbf{0.3272} \\
    \bottomrule
    \end{tabular}
    }
    \label{table_magic_of_time}
\end{table}

\begin{figure*}[!t]
\centering
\begin{subfigure}{0.44\textwidth}
  \centering
  \includegraphics[width=\linewidth]{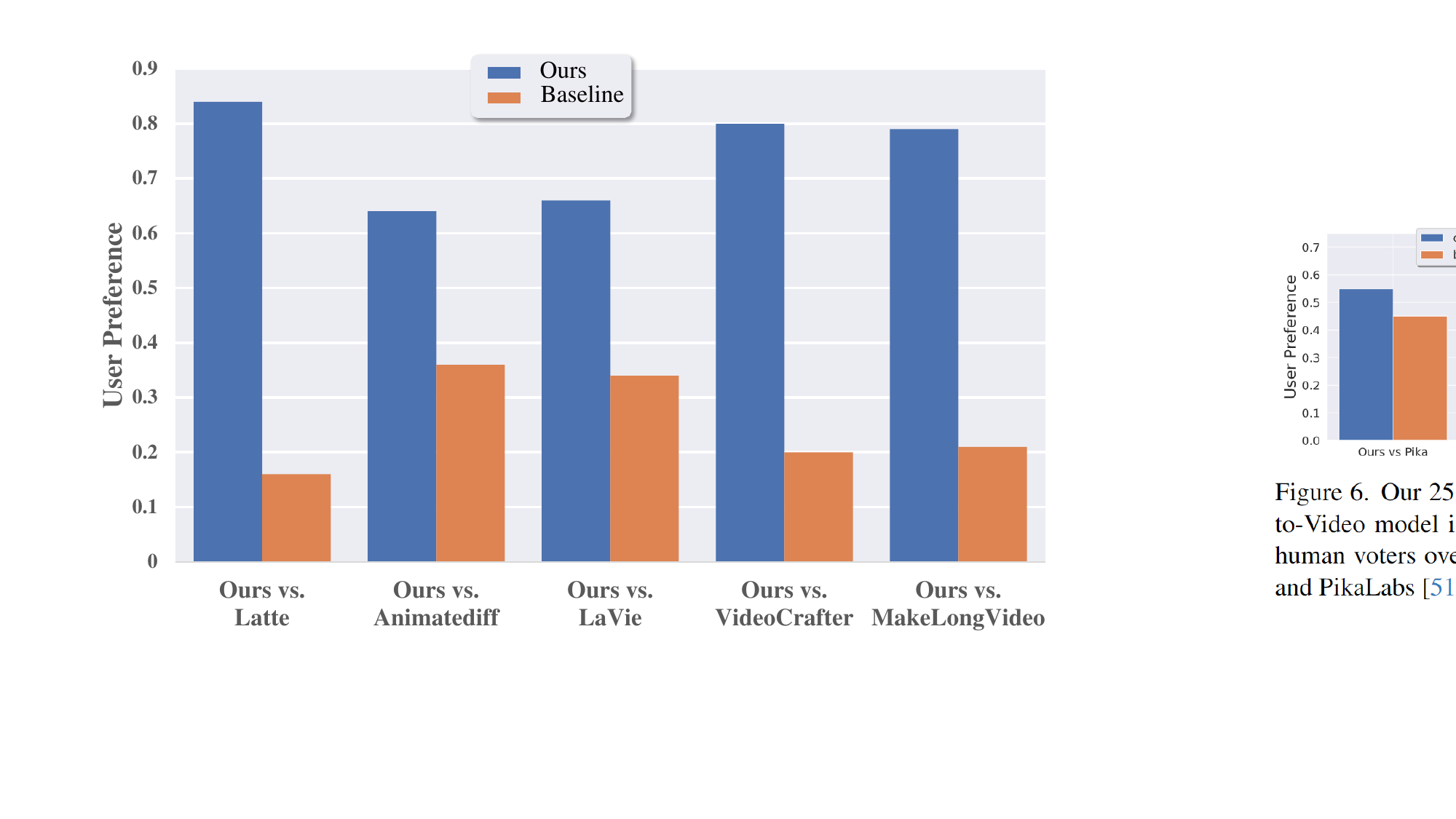}
  \caption{}
  \label{figure_user_comparision}
\end{subfigure}%
\hspace{0.05\textwidth} 
\begin{subfigure}{0.44\textwidth}
  \centering
  \includegraphics[width=\linewidth]{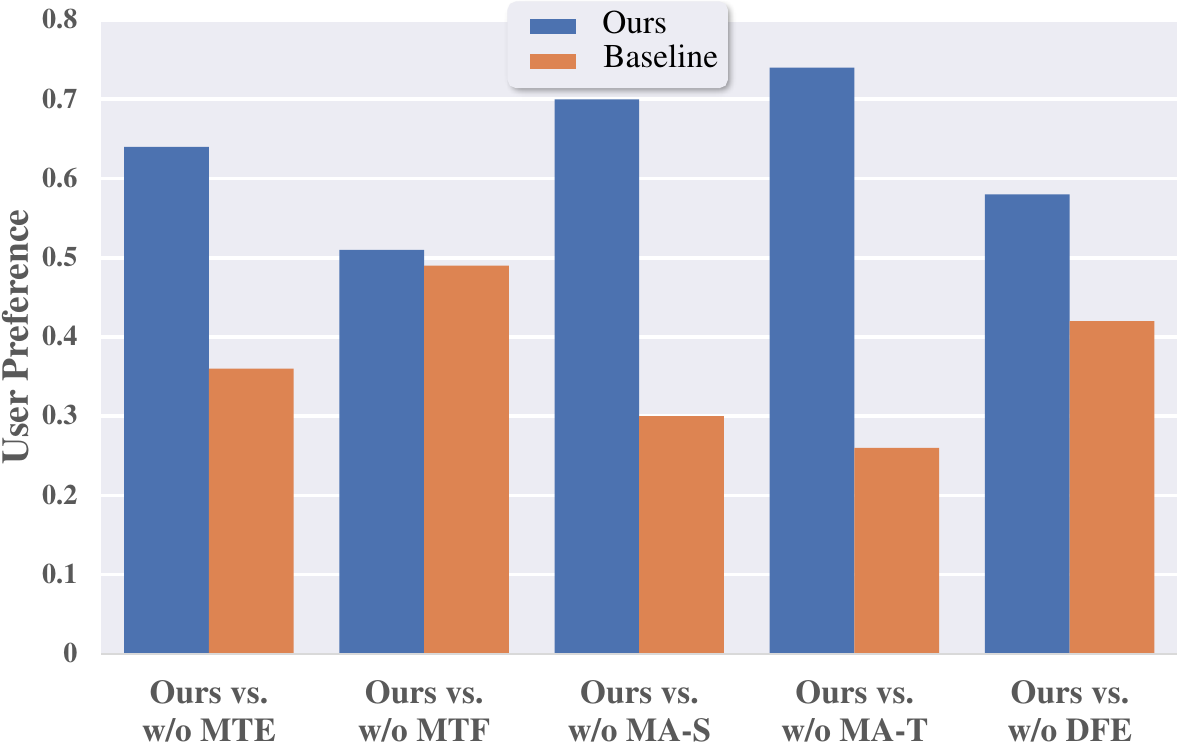}
  \caption{}
  \label{figure_user_ablation}
\end{subfigure}
\caption{\textbf{(a) Human evaluation between MagicTime and state-of-the-art T2V generation methods.} MagicTime is preferred by human voters over other methods. \textbf{(b) Human evaluation of different ablation settings.} The most crucial components are \textit{MagicAdapter-S/T} (MA-S/T) and \textit{Dynamic Frames Extraction} (DFE), which empower MagicTime with metamorphic video generation capabilities. \textit{Magic Text-Encoder} (MTE) and \textit{Multi-view Text Fusion} (MTF) are used to enhance video quality.}
\label{figure_both}
\end{figure*}

\begin{figure*}[t]
  \centering
  \includegraphics[width=0.954\linewidth]{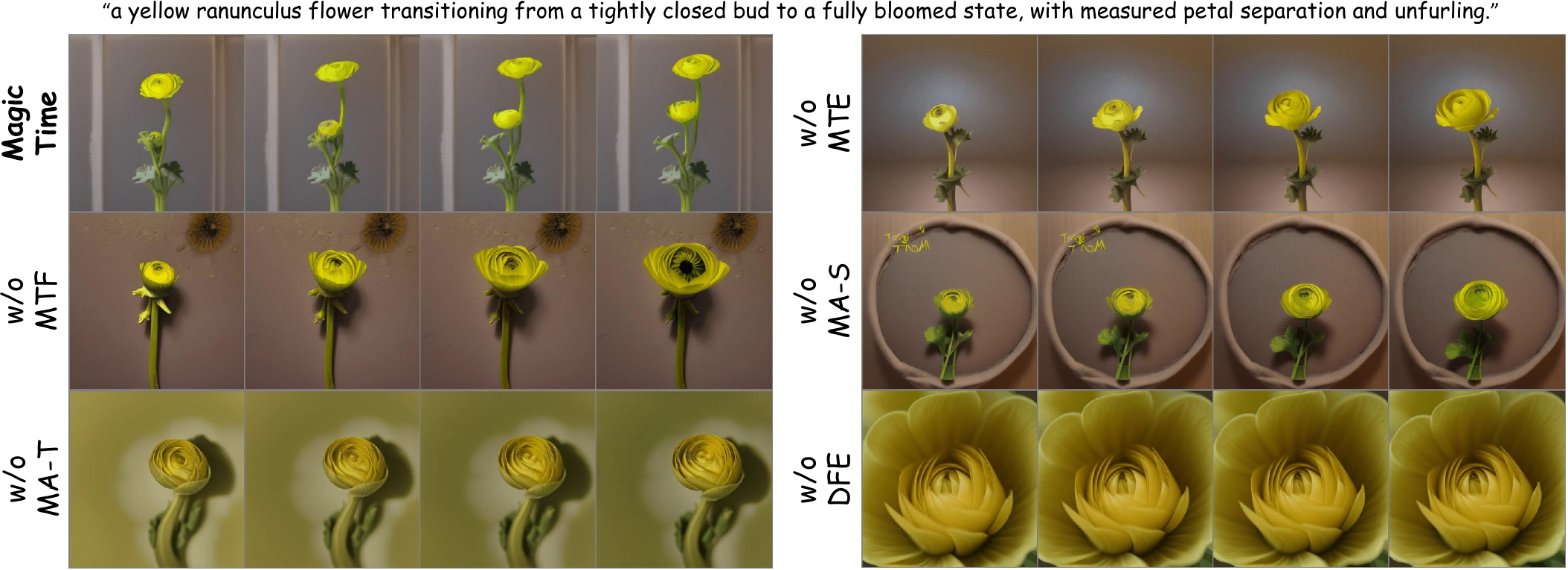}
  \caption{\textbf{Ablation study for different parts of MagicTime.} It shows the effectiveness of \textit{Magic Text-Encoder} (MTE), \textit{Multi-view Text Fusion} (MTF), \textit{MagicAdapter-S/T} (MA-S/T), and \textit{Dynamic Frames Extraction} (DFE).
  }
  \label{figure_ablation_study}
\end{figure*}

\subsection{Qualitative Analysis}

In this section, we compare our method, MagicTime, with several state-of-the-art models including MakeLongVideo \cite{MakeLongVideo}, ModelScopeT2V \cite{Modelscope}, VideoCrafter2 \cite{VideoCrafter2}, Zeroscope \cite{zeroscope}, LaVie \cite{lavie}, T2V-zero \cite{T2V-zero}, Latte \cite{Latte}, and Animatediff \cite{animatediff} in the task of metamorphic video generation. All input text is sourced from our ChronoMagic dataset. The evaluated tasks involve generating videos that depict significant metamorphosis, such as the construction of a modern house in Minecraft, the blooming of pink plum blossoms, the baking of bread rolls, and the melting of ice cubes, as shown in Fig. \ref{figure_comparision_CTL_1}. Our evaluation focuses on Frame Consistency (FC), Metamorphic Amplitude (MA), and Text Alignment (TA). While all methods, except MakeLongVideo, produce high-quality videos with strong temporal modeling implying high FC, they struggle with TA and MA. The limited physical priors in existing text-to-video models restrict them to general scene generation, such as swaying flowers (ModelScopeT2V, VideoCrafter2, Latte), static buildings (Zeroscope, LaVie, Animatediff, T2V-zero), which are incorrect contents. In contrast, MagicTime excels in capturing metamorphic transitions smoothly and accurately reflecting physical sequences, such as layer-by-layer Minecraft construction and natural flower blooming. Notably, MagicTime demonstrates the ability to generate visually consistent content over large time spans, accurately depicting processes like ice melting and dough baking, further validating the model's effectiveness. Please refer to the Appendix for more details.

\subsection{Quantitative Analysis}
In this section, we conduct a comprehensive quantitative evaluation to assess the performance of MagicTime. We compare the generation quality on the ChronoMagic dataset between MagicTime and baseline methods, as presented in Table \ref{table_magic_of_time}. Consistent with the results in Fig. \ref{figure_comparision_CTL_1}, MakeLongVideo exhibits the highest FID and FVD values, indicating its inferior visual quality compared to other methods. While VideoCrafter2 has the lowest FID, our method boasts the lowest FVD, suggesting that MagicTime offers the highest video quality. CLIPSIM, originally applied in image generation, mainly measures the similarity between individual video frames and text for videos, which limits its effectiveness in assessing video generation quality. Given the lack of superior metrics in the T2V field, we use CLIPSIM as an indicator of video-text relevance. Numerically, although the video-text relevance of MagicTime is the highest, it is only slightly higher than the other methods. However, as demonstrated in Fig. \ref{figure_comparision_CTL_1}, only MagicTime adheres to the text prompt instructions to generate metamorphic videos, while the baseline methods generate general videos based on the theme of the prompt, implying that their CLIPSIM scores should be lower. Overall, our method surpasses all compared methods, confirming that in modeling metamorphic videos, we can ensure superior video generation quality by successfully incorporating more physical priors.

\subsection{Human Evaluation}\label{sec:human_eval}
As previously mentioned, metrics such as FID \cite{FID}, FVD \cite{FVD}, and CLIPSIM \cite{CLIP} may not fully capture the generative capabilities of T2V methods. To address this, we enlisted 200 participants to evaluate the Visual Quality, Frame Consistency, Metamorphic Amplitude, and Text Alignment of videos generated by MagicTime and baseline methods. We selected 15 sets of samples for each criterion and asked participants to choose the high-quality metamorphic video that best aligned with their expectations. For fairness, all options were randomized, and each participant could only complete one questionnaire under identical conditions. Due to the need for a large pool of participants in the evaluation, we only select the top 5 T2V methods for voting. We finally collected 185 valid questionnaires. The results in Fig. \ref{figure_user_comparision} demonstrate that our method significantly outperforms other leading T2V methods. Furthermore, MagicTime significantly enhances the quality of metamorphic videos, confirming the effectiveness of encoding physical knowledge through training with metamorphic videos. Visualizations of the questionnaire are provided in the Appendix.

\begin{figure*}[t]
	\centering
	\includegraphics[width=0.923\linewidth]{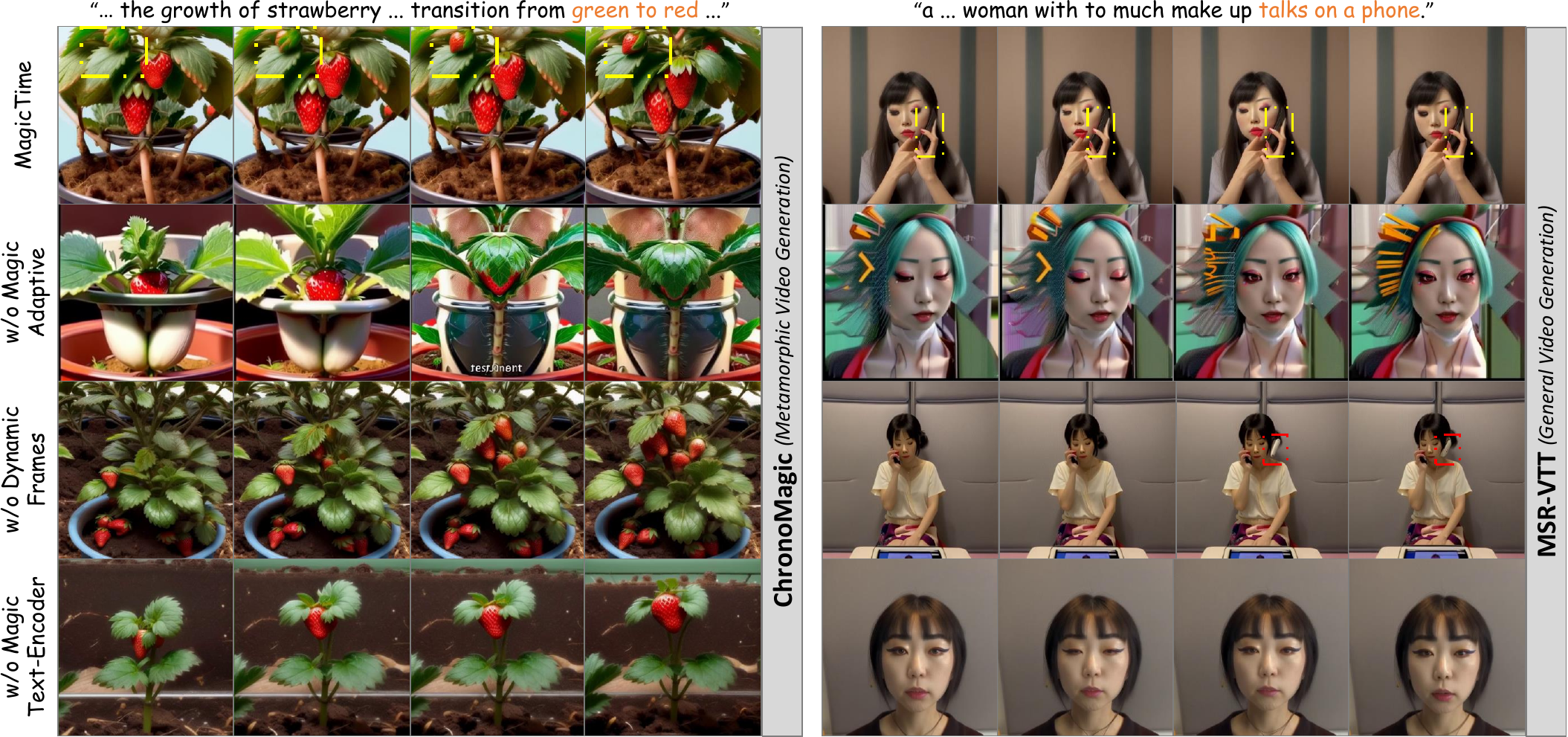}
	\caption{\textbf{Further Validation on \textit{Magic Adaptive Strategies}, \textit{Dynamic Frame Extraction} and \textit{Magic Text-Encoder}.} Only by adopting the above three methods can we efficiently encode the metamorphic prior without losing the general video generation capability. The \textcolor{yellow!95!red}{yellow dashed box} corresponds to the content of \textcolor{orange!88!red}{orange word}.}
    \label{figure_ablation_on_general_forgetting_1}
\end{figure*}

\subsection{Ablation Study}
\myparagraph{Validation on Different Components}
We primarily conduct ablation studies on our proposed strategies of Dynamic Frames Extraction (DFE), MagicAdapter-T (MA-T), MagicAdapter-S (MA-S), Multi-view Text Fusion (MTF) and Magic Text-Encoder (MTE), as illustrated in Fig. \ref{figure_ablation_study}. Given the prompt "A bud transforms into a yellow flower", the Group w/o MTE, can produce relatively better results. However, compared to the complete MagicTime, it still slightly lags in aspects such as fluency and degree of metamorphosis. The Group w/o MTF, due to the use of low-quality prompts, resulted in a decrease in generation quality, such as the appearance of strange buds in the middle of flowers. The Group w/o MA-S can generate the complete process, but as the dataset contains a large amount of watermark data, the results also include watermarks. The Group w/o MA-T struggles to effectively encode the concept of metamorphic, resulting in general video without metamorphic characteristics. The Group w/o DFE (use \textit{random frame extraction} instead), similarly to the Group w/o MA-T, can not generate metamorphic video and has reduced generation quality due to the lack of metamorphic feature acquisition. We also conduct quantitative experiments, as shown in Fig. \ref{figure_user_ablation}, demonstrating that MagicTime significantly outperforms other methods across all four dimensions. The experimental settings are consistent with those in Section \ref{sec:human_eval}, and we finally collected 115 valid questionnaires. Our MagicTime significantly enhances metamorphic video generation. With simple adapters and training strategies, it validates the hypothesis that incorporating time-lapse videos into the model can encode more physical knowledge.



\noindent\myparagraph{Validation on Training Scheme}
To illustrate the gains of the proposed MagicTime Recipe, we evaluate based on different ablation experiment groups using ChronoMagic (metamorphic videos) and perform zero-shot inference on 1K prompts which are randomly sampled from MSR-VTT~\cite{MSR-VTT} (general videos). The results are shown in Fig. \ref{figure_ablation_on_general_forgetting_1}. MagicTime consistently achieves superior visual quality and text-following capability. Specifically, the "\textit{w/o Magic Adaptive Strategy}" (joint training MagicAdapter S \& T instead) shows severe flickering (e.g., strawberry) and strange green edges around the video (e.g., strawberry, woman). We speculate that this is due to the significant difference between the metamorphic process and the general process. The former covers the complete transformation process of objects, while the latter primarily includes camera movements, and a general training strategy cannot enable the model to distinguish between them. Only the decoupled training strategy can effectively learn the metamorphic process by encoding the physical priors, rather than by fitting the training data, so that it can learn to generate metamorphic videos without losing the original ability. Intuitively, uniform sampling for all metamorphic videos seems optimal. However, the "\textit{w/o Dynamic Frame Extraction}" (use uniform frame extraction instead) proves this comes at the expense of general video generation capability, meaning the model learns the metamorphic process by fitting data rather than encoding physical information, resulting in erratic changes in generated general videos (e.g., phones suddenly appearing on face). The \textit{Magic Text Encoder} demonstrates a refined understanding of prompts. For example, only "\textit{MagicTime}" can accurately generate the yellow box content as required by the orange text, while other groups can only understand coarse-grained information. We also provide quantitative analysis results, which are generally similar to the qualitative results, as shown in Table \ref{table_ablation_on_training_strategy}. Due to the inaccuracy of existing automatic evaluation metrics, the results are for reference only. More results can be found in Appendix.


\begin{table}[t]
    \centering
    \caption{\textbf{Validation on \textit{Magic Adaptive Strategies}, \textit{Dynamic Frame Extraction} and \textit{Magic Text-Encoder} by Automatic Metrics.} Removing any of the above methods significantly reduces the video's visual quality and the model's ability to follow instructions.}
    \label{table_ablation_on_training_strategy}
    \begin{tabular}{lccc}
    \toprule
    \textbf{Method} & \textbf{FID$\downarrow$} & \textbf{FVD$\downarrow$} & \textbf{CLIPSIM$\uparrow$} \\
    \midrule
    \multicolumn{4}{c}{\textbf{ChronoMagic} (metamorphic)} \\
    \midrule
    \rowcolor{myblue} MagicTime & \textbf{102.91} & 449.92 & \textbf{0.3230} \\
    w/o Magic Adaptive & 106.91 & 506.54 & 0.3173 \\
    w/o Dynamic Frames & 110.16 & 462.33 & 0.3142 \\
    w/o Magic Text-Encoder & 108.87 & \textbf{432.58} & 0.3073 \\
    \midrule
    \multicolumn{4}{c}{\textbf{MSR-VTT} (general)} \\
    \midrule
    \rowcolor{myblue} MagicTime & 105.06 & \textbf{523.87} & 0.2775 \\
    w/o Magic Adaptive & \textbf{87.29} & 543.55 & 0.2734 \\
    w/o Dynamic Frames & 130.14 & 555.76 & 0.2767 \\
    w/o Magic Text-Encoder & 104.53 & 544.31 & \textbf{0.2815} \\
    \bottomrule
    \end{tabular}%
\end{table}


\section{Conclusion}
In this paper, we present MagicTime, a novel framework for metamorphic time-lapse video generation. It can seamlessly integrate into existing community models, and generate metamorphic videos conforming to the content distribution and motion patterns in real-captured metamorphic videos. Extensive experiments demonstrate our effectiveness in generating high-quality metamorphic videos with various prompts and styles. Further, our MagicTime model and ChronoMagic benchmark can provide a simple yet effective solution for metamorphic video generation, which can facilitate further investigation by the community. 


\clearpage
\newpage
\bibliographystyle{IEEEtran}
\bibliography{refs-clean}

\newpage

\begin{IEEEbiography}[{\includegraphics[width=1in,height=1.25in,clip,keepaspectratio]{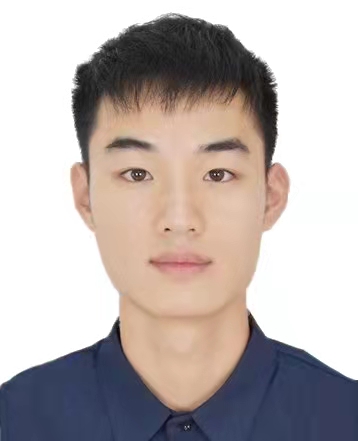}}]{Shenghai Yuan} is a master's student in the School of Electronic and Computer Engineering with Peking University (PKU), advised by Prof. Li Yuan. His research interests include but are not limited to multimodal understanding and multimodal generation. He has gradually expanded to video generation models, and his notable works include Open-Sora Plan, MagicTime, ConsisID, and ChronoMagic-Bench.
\end{IEEEbiography}

\begin{IEEEbiography}[{\includegraphics[width=1in,height=1.25in,clip,keepaspectratio]{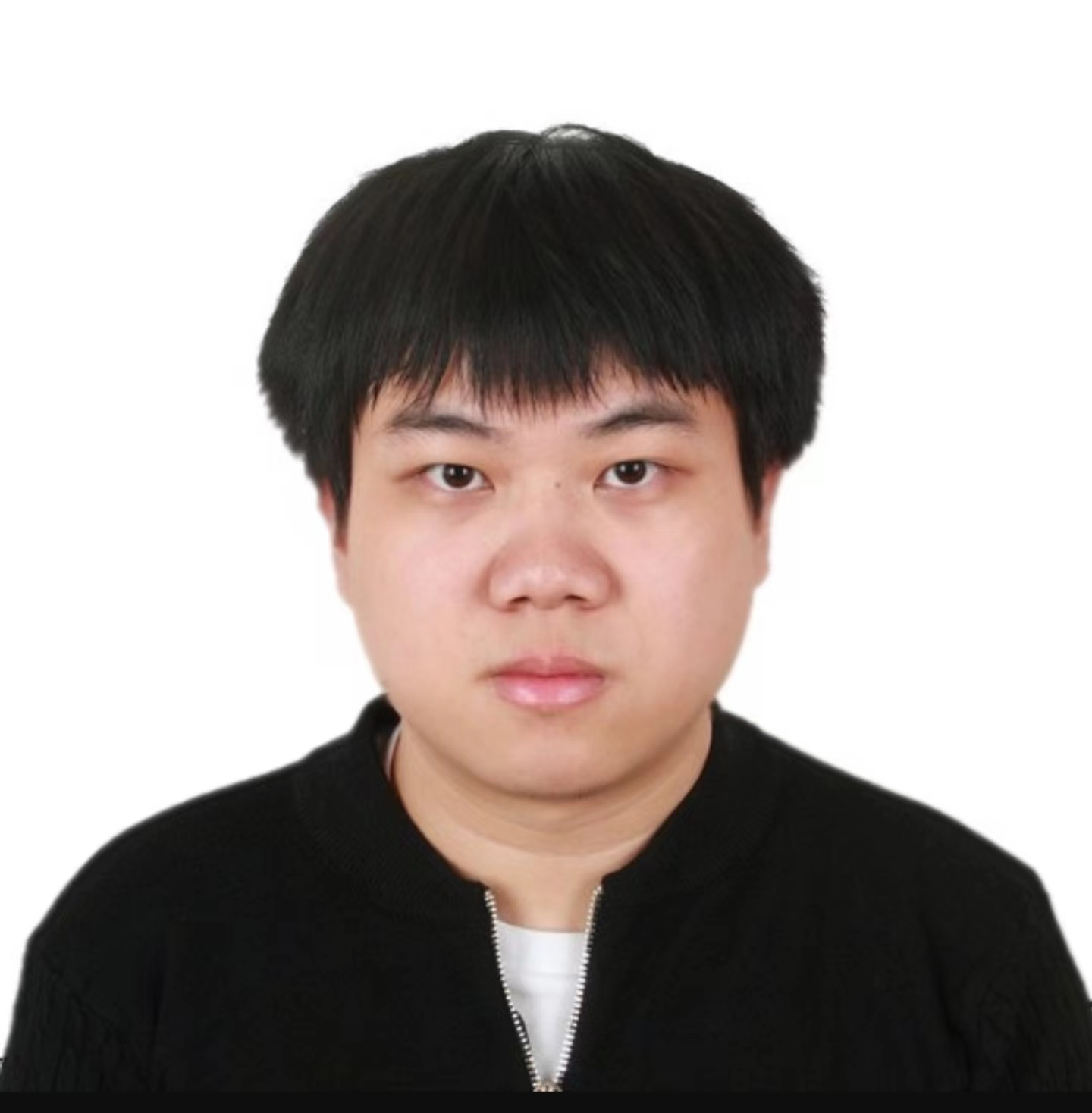}}]{Jinfa Huang} is a Ph.D. student in the Department of Computer Science, University of Rochester (UR), advised by Prof. Jiebo Luo. He aims at building multimodal interactive AI systems that can not only ground and reason over the external world signals. As steps towards this goal, His research interests include but are not limited to multimodal understanding and multimodal generation. Prior to that, he got his master's degree from Peking University (PKU) in 2023, advised Prof. Jie Chen. 
\end{IEEEbiography}

\begin{IEEEbiography}[{\includegraphics[width=1in,height=1.25in,clip,keepaspectratio]{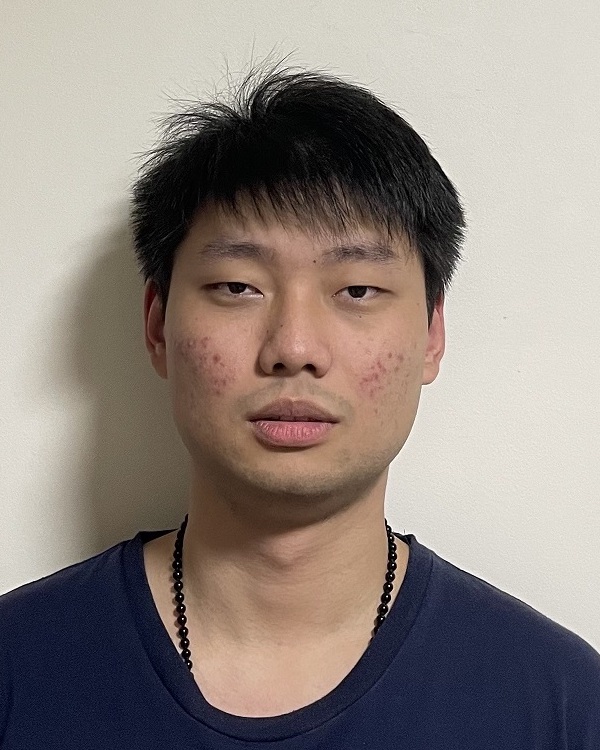}}]{Yunjun Shi} is currently a PhD student in the Department of Electrical and Computer Engineering, National University of Singapore, advised by Prof. Vincent Y.~F.~Tan. He received a B.Eng. in Computer Science from Nankai University in 2019. His research interests focus on generative image editing, as well as tackling distribution shift scenarios in deep learning such as Continual Learning, Federated Learning, etc. 
\end{IEEEbiography}

\begin{IEEEbiography}[{\includegraphics[width=1in,height=1.25in,clip,keepaspectratio]{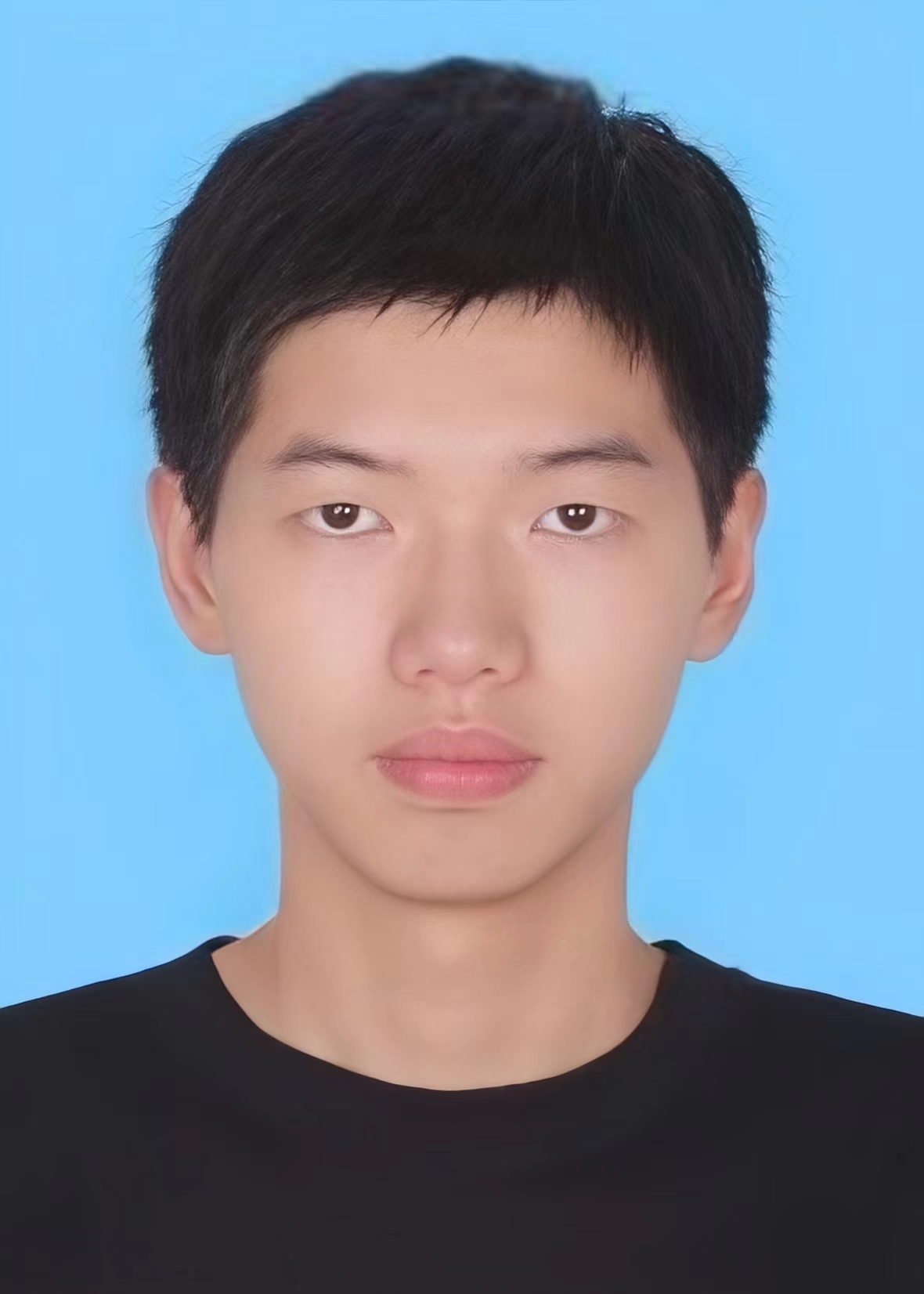}}]{Yongqi Xu} is with the School of Computer Science and Technology, Guangdong University of Technology, Guangzhou, China. He is a master's candidate at the School of Environment and Energy, Peking University (PKU). His current research interests include Computer Vision and Artificial Intelligence Generated Content.
\end{IEEEbiography}

\begin{IEEEbiography}[{\includegraphics[width=1in,height=1.25in,clip,keepaspectratio]{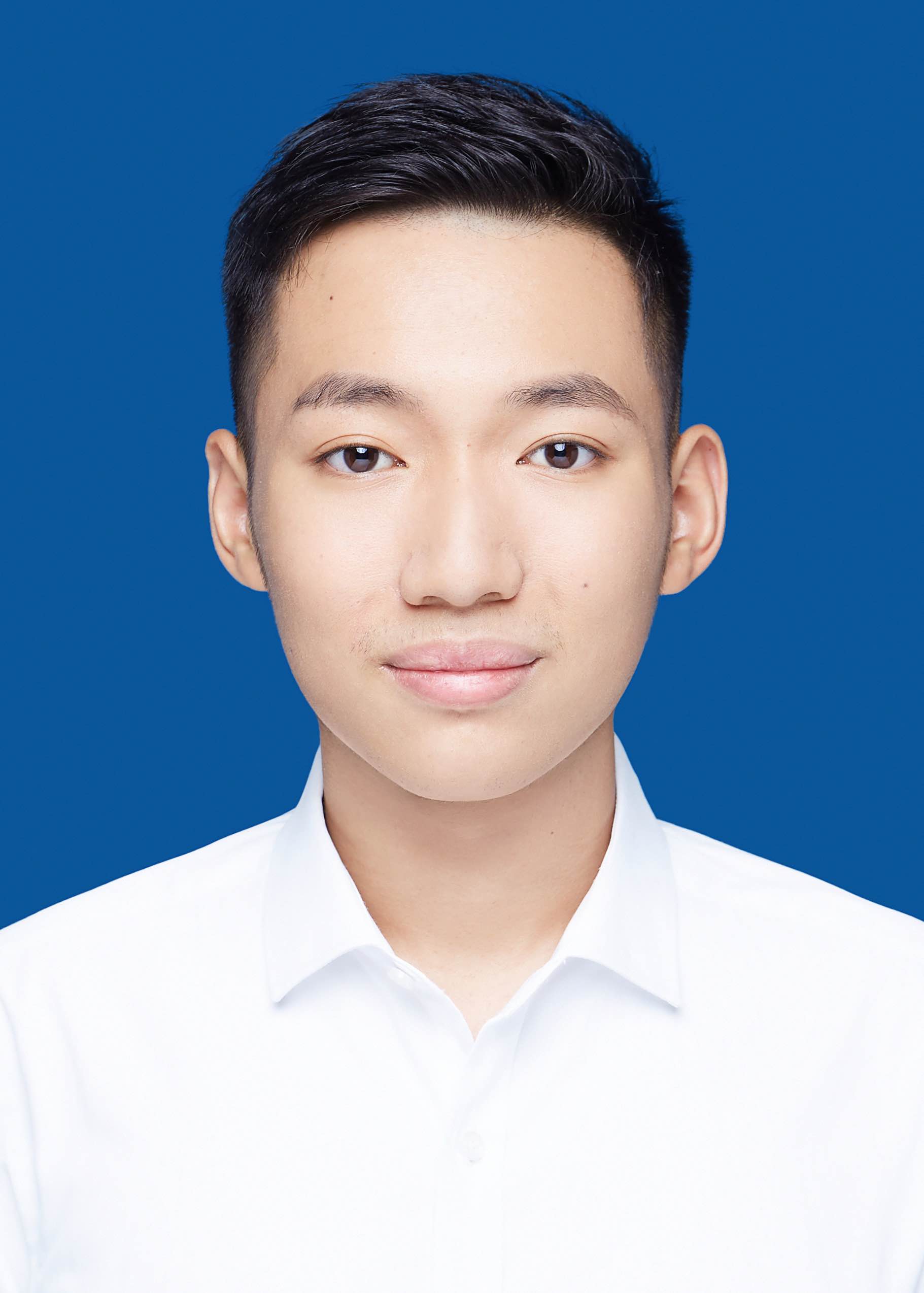}}]{Ruijie Zhu} is a Ph.D. student in the Department of Electrical and Computer Engineering at the University of California, Santa Cruz (UCSC), advised by Prof. Jason K. Eshraghian, having commenced his program in Fall 2023. His research centers on scaling up models with a focus on efficient inference, integrating deep learning with neuromorphic computing. His initial research focus was on spiking neural networks, and during his undergraduate studies, he contributed to multiple open-source neuromorphic projects. Prior to joining UCSC, he received his Bachelor's degree from the University of Electronic Science and Technology of China (UESTC) in 2023, where he was advised by Prof. Liang-Jian Deng. 
\end{IEEEbiography}

\begin{IEEEbiography}[{\includegraphics[width=1in,height=1.25in,clip,keepaspectratio]{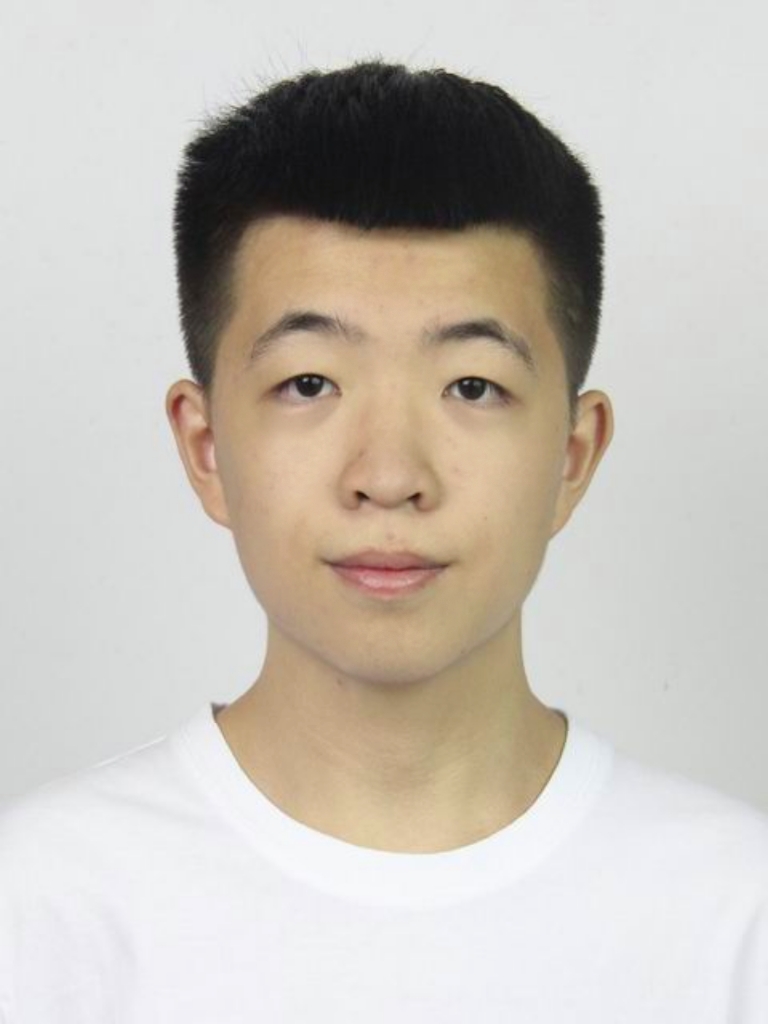}}]{Lin Bin} is a master's student at Peking University, specializing in Computer Application Technology under the supervision of Yuan Li. He primarily focuses on multimodal large models, video understanding, and generation. Initially targeting visual understanding, he has gradually expanded to video generation models and is now delving into multimodal generation and understanding tasks. His notable works include Open-Sora Plan, Video-LLaVA, and MoE-LLaVA.
\end{IEEEbiography}

\begin{IEEEbiography}[{\includegraphics[width=1in,height=1.25in,clip,keepaspectratio]{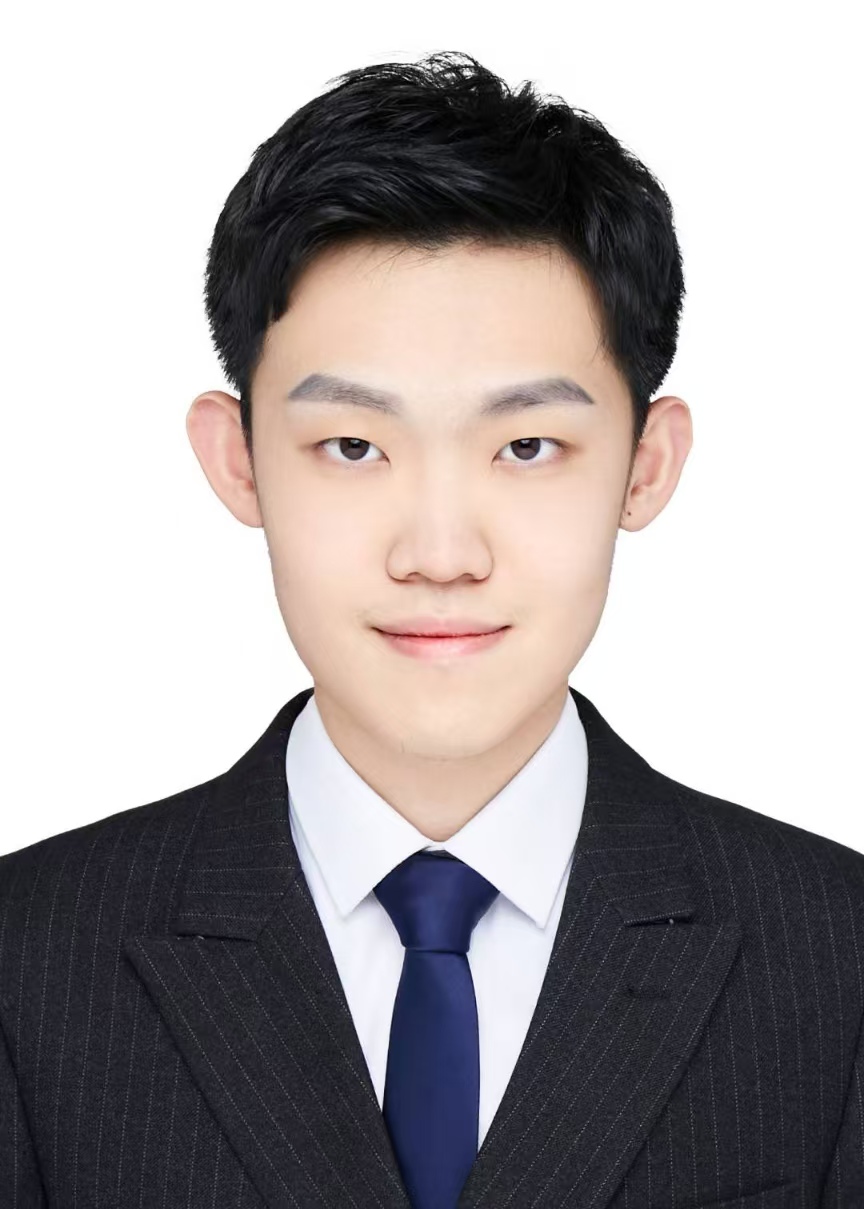}}]{Xinhua Cheng} received the B.E. degree in Computer Science and Technology from the College of Computer Science, Sichuan University, Chengdu, China. He is currently pursuing a Ph.D degree at the School of Electronic and Computer Engineering, Peking University, Shenzhen, China. His recent research interests include Artificial Intelligence Generated Content, especially video and 3D content generation.
\end{IEEEbiography}

\begin{IEEEbiography}[{\includegraphics[width=1in,height=1.25in,clip,keepaspectratio]{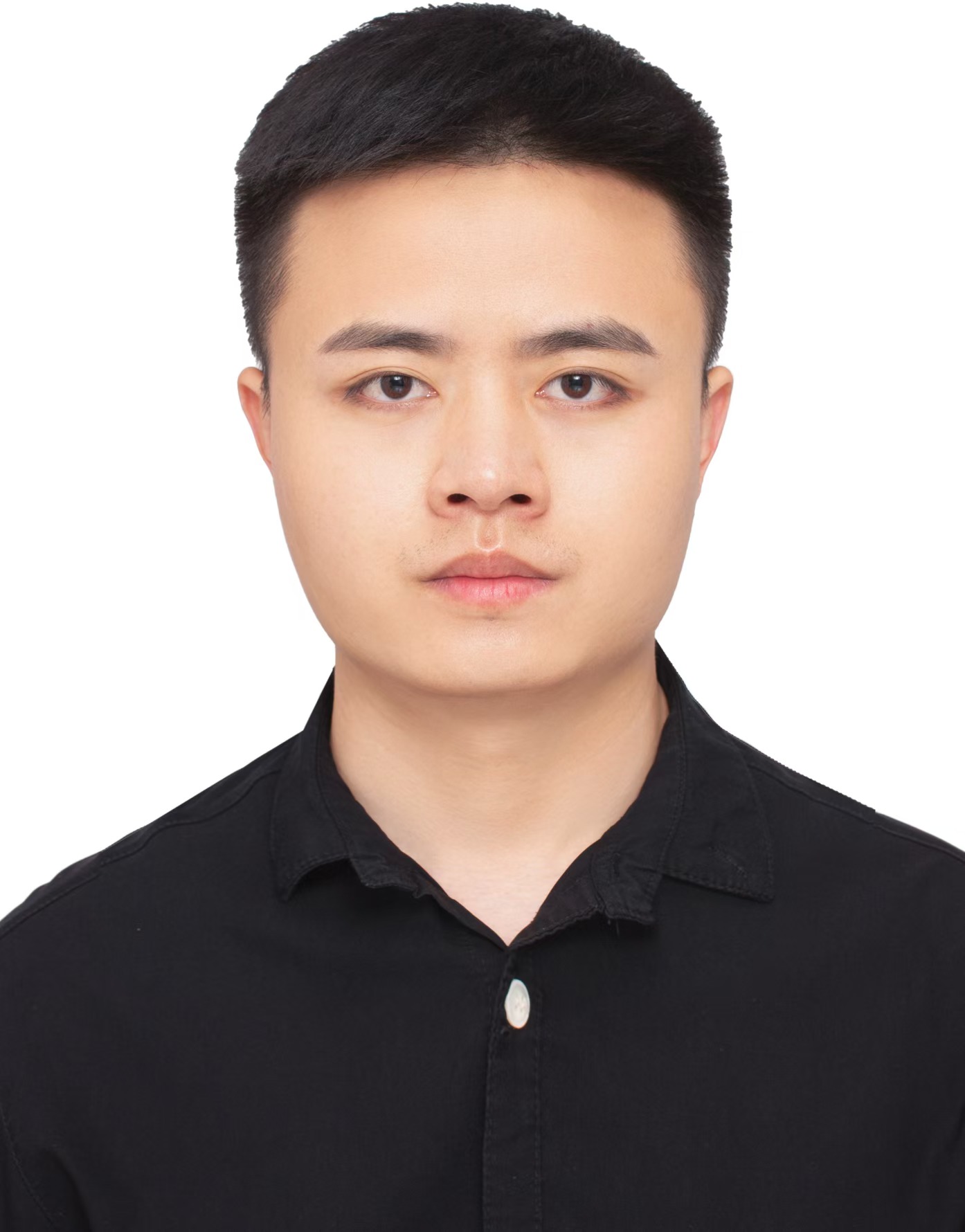}}]{Li Yuan} received the B.Eng. degree from the University of Science and Technology of China, Hefei, China, in 2017, and the Ph.D. degree from the National University of Singapore, Singapore, in 2021. He is currently a tenure-track Assistant Professor with the School of Electrical and Computer Engineering, Peking University, the Peng Cheng Laboratory, Shenzhen, China. He has authored or co-authored more than 100 journals/conference papers in computer vision and machine learning. His research interests include computer vision and deep learning.
\end{IEEEbiography}

\begin{IEEEbiography}[{\includegraphics[width=1in,height=1.25in,clip,keepaspectratio]{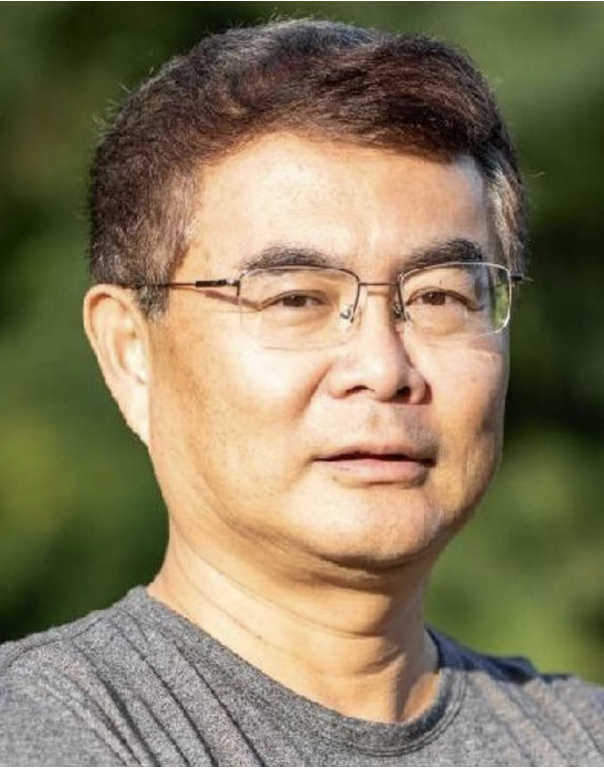}}]{Jiebo Luo} is a professor of computer science at the University of Rochester, which he joined in 2011 after a prolific career of fifteen years at Kodak Research Laboratories. He has authored nearly 600 technical papers and holds more than 90 U.S. patents. His research interests include computer vision, NLP, machine learning, data mining, computational social science, and digital health. He has served on the editorial boards of IEEE Transactions on Pattern Analysis and Machine Intelligence, IEEE Transactions on Multimedia, IEEE Transactions on Circuits and Systems for Video Technology, IEEE Transactions on Big Data, ACM Transactions on Intelligent Systems and Technology, and Pattern Recognition. He was the EIC of the IEEE TMM from 2020-2022. He is a fellow of ACM, AAAI, IEEE, SPIE, and IAPR.
\end{IEEEbiography}


\clearpage
\newpage
\appendices
\setcounter{page}{1}

\startcontents[chapters]
\printcontents[chapters]{}{1}{}

\section{Additional Experimental Results}\label{sec: Additional Experimental Results}

\begin{figure}[!t]
	\centering
	\includegraphics[width=1\linewidth]{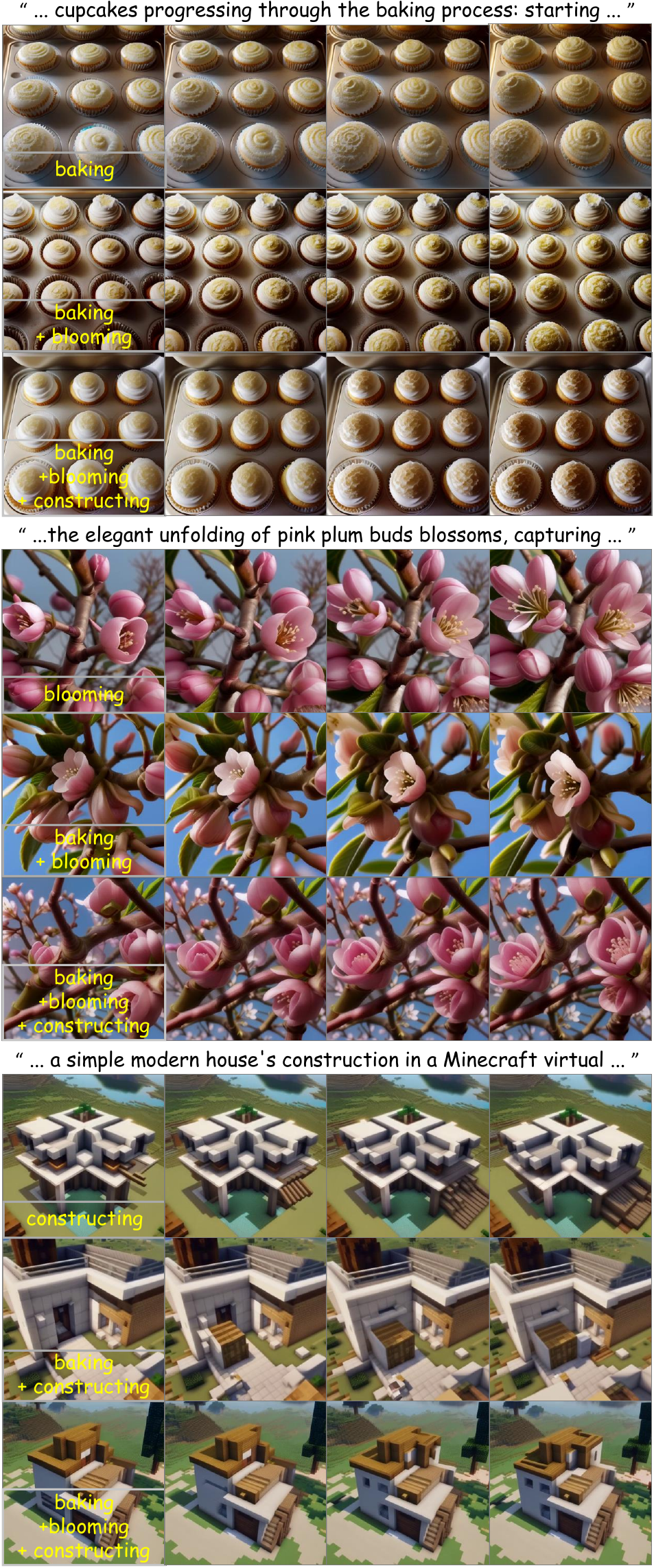}
	\caption{\textbf{Incremental Ablation Study on Training Data Categories.} As the number of training tasks increases, the video quality improves, indicating mutual gain among the tasks.}
	\label{figure_ablation_on_dataset_forgetting}
\end{figure}

\subsection{Validation on Catastrophic Forgetting}
\begin{figure*}[!t]
	\centering
	\includegraphics[width=1\linewidth]{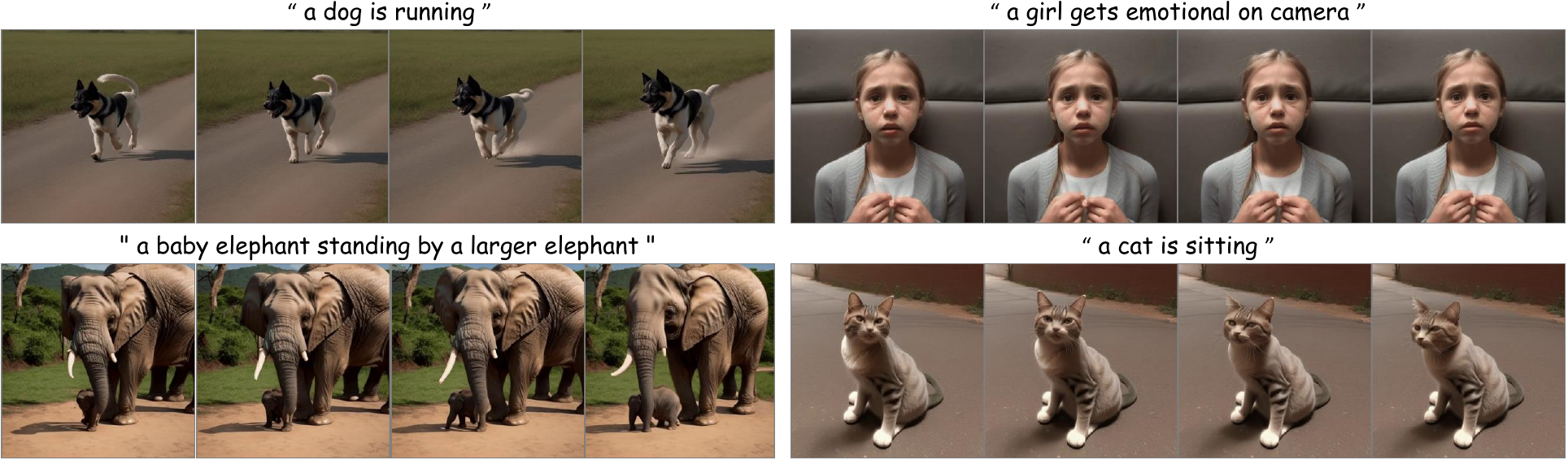}
	\caption{\textbf{Qualitative experiment of our MagicTime on general video generation.} We perform zero-shot inference on MSR-VTT~\cite{MSR-VTT}, and MagicTime is still able to generate high-quality general videos.}
	\label{figure_ablation_MSRVTT}
\end{figure*}

We first verify whether catastrophic forgetting exists when training on a vast different set of videos involving different kinds of metamorphic physical processes. Due to the wide range of video categories in ChronoMagic, we select videos from the baking, blooming, and constructing categories, which involve distinctly different types of motion, for verification. We conduct incremental ablation studies on these dataset categories, based on the complete training process of MagicTime, and adjust the training epochs according to the amount of data. The results are shown in Fig. \ref{figure_ablation_on_dataset_forgetting}. Intuitively, multi-task learning is more challenging than single-task learning, as the former involves focusing on a greater number of task categories. However, our experimental results indicate that there is no conflict between different tasks; instead, as the number of training tasks increases, they promote each other. For instance, in the \textit{first line} (e.g. \textit{one task learning}), the lighting on the lower section of the cupcakes is uneven, being brighter on the left and darker on the right, whereas the top is uniformly bright. Additionally, the color of the bread does not scorch naturally over time. Conversely, in the \textit{second} and \textit{third lines} (e.g. \textit{multi-task learning}), these issues are absent; the \textit{fourth} and \textit{fifth lines} demonstrate that both the visual quality and the metamorphic process of the flowers and buildings improve and appear more natural as the number of training tasks increases. We speculate that in addition to the gain between training data, our MagicTime Recipe can effectively take advantage of the richness of training categories. Then, we verify whether catastrophic forgetting exists for general video generation after incorporating metamorphic physics priors. The experiment is based on MSR-VTT~\cite{MSR-VTT}, and the results are shown in Fig. \ref{figure_ablation_MSRVTT} and Fig. \ref{figure_ablation_on_general_forgetting_2}. It can be observed that although MagicTime focuses on metamorphic video generation, its performance in general video generation is not compromised. This demonstrates that we can preserve original motion priors through the \textit{Magic Adaptive Strategy}, \textit{Dynamic Frames Extraction}, and \textit{Magic Text-Encoder}, while simultaneously encoding additional metamorphic priors.

\subsection{Validation of Out-of-Distribution Dataset}
\begin{figure}[!t]
    \centering
    \includegraphics[width=1\linewidth]{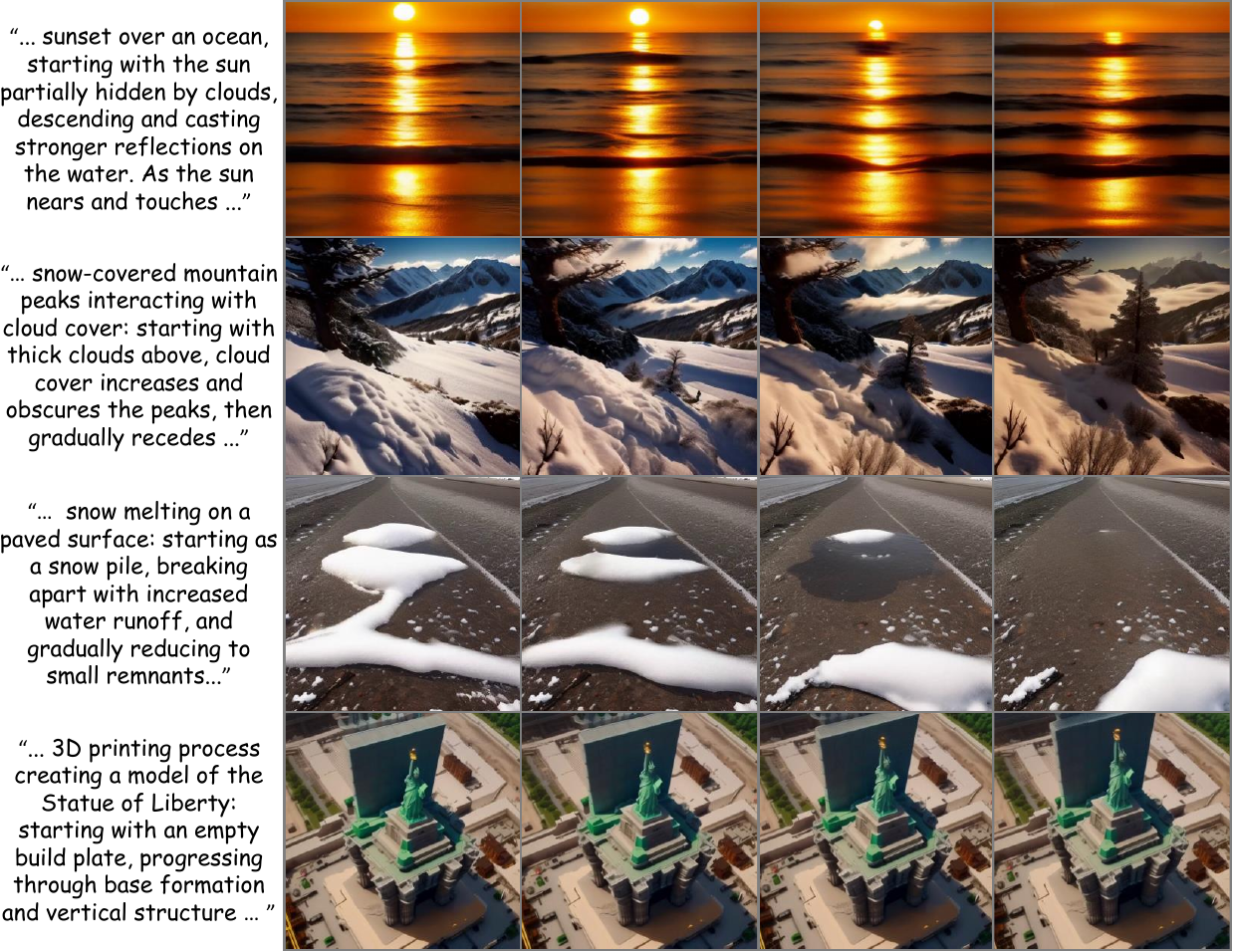}
    \caption{\textbf{Out-of-Distribution Concepts Generation.} MagicTime can generalize to some simple time-lapses (e.g., sunset, seasonal changes, snow melt), but it struggles to generate complex time-lapse processes (e.g., 3D printing).}
    \label{figure_Out-of-Distribution Concepts}
\end{figure}

\jf{
To further evaluate robustness, we tested several samples that are completely out-of-distribution concepts of ChronoMagic (e.g., bloom, seeds germinating, building construction, baking). Fig. \ref{figure_Out-of-Distribution Concepts} clearly illustrates that MagicTime is adept at generalizing simple time-lapse processes such as Case 1’s sunset, Case 2’s seasonal changes, and Case 3’s snow melt. However, in more complex scenarios like Case 4's 3D printing, MagicTime fails to produce the entire time-lapse following prompt but can improve the motion range.
}

\subsection{Validation on DiT-based Models}
\begin{figure}[t]
	\centering
	\includegraphics[width=1\linewidth]{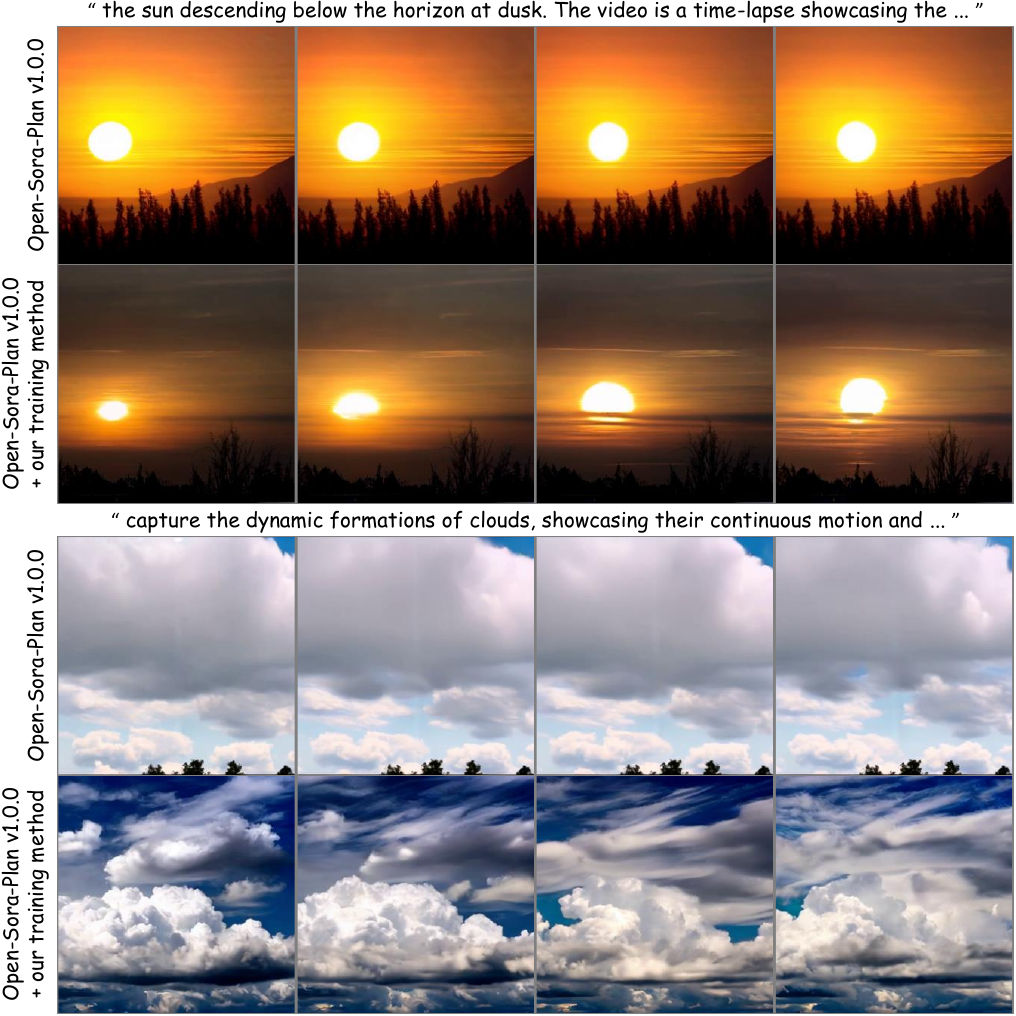}
	\caption{\textbf{Validation of MagicTime Training Recipe on Open-Sora-Plan v1.0.0.} After training, the baseline can generate landscape metamorphic videos, whereas previously it could only produce relatively static videos.}
	\label{figure_opensoraplan_ijcv}
\end{figure}

DiT-based Sora~\cite{SORA} demonstrates significant potential in simulating the physical world. To further verify the effectiveness of the proposed MagicTime Recipe (e.g., \textit{Magic Adaptive Strategy}, \textit{Dynamic Frame Extraction}, and \textit{Magic Text Encoder}), we then integrate it into the DiT-based framework~\cite{Latte, opensora, opensoraplan}. We select Open-Sora-Plan v1.0.0~\cite{opensoraplan} as a representative, given its superior performance in generating landscape videos. We then expand the dataset with additional metamorphic landscape time-lapse videos using the same annotation framework to create the ChronoMagic-Landscape dataset. Subsequently, we train Open-Sora-Plan v1.0.0 with this dataset and our MagicTime Recipe. As illustrated in Fig. \ref{figure_opensoraplan_ijcv}, our model can generate high-quality metamorphic landscape time-lapse videos, whereas the baseline model produces nearly static videos. This further validates the effectiveness of our MagicTime Recipe.

\subsection{Validation on Large Video Generative Models}
\begin{figure}[!t]
	\centering
	\includegraphics[width=0.95\linewidth]{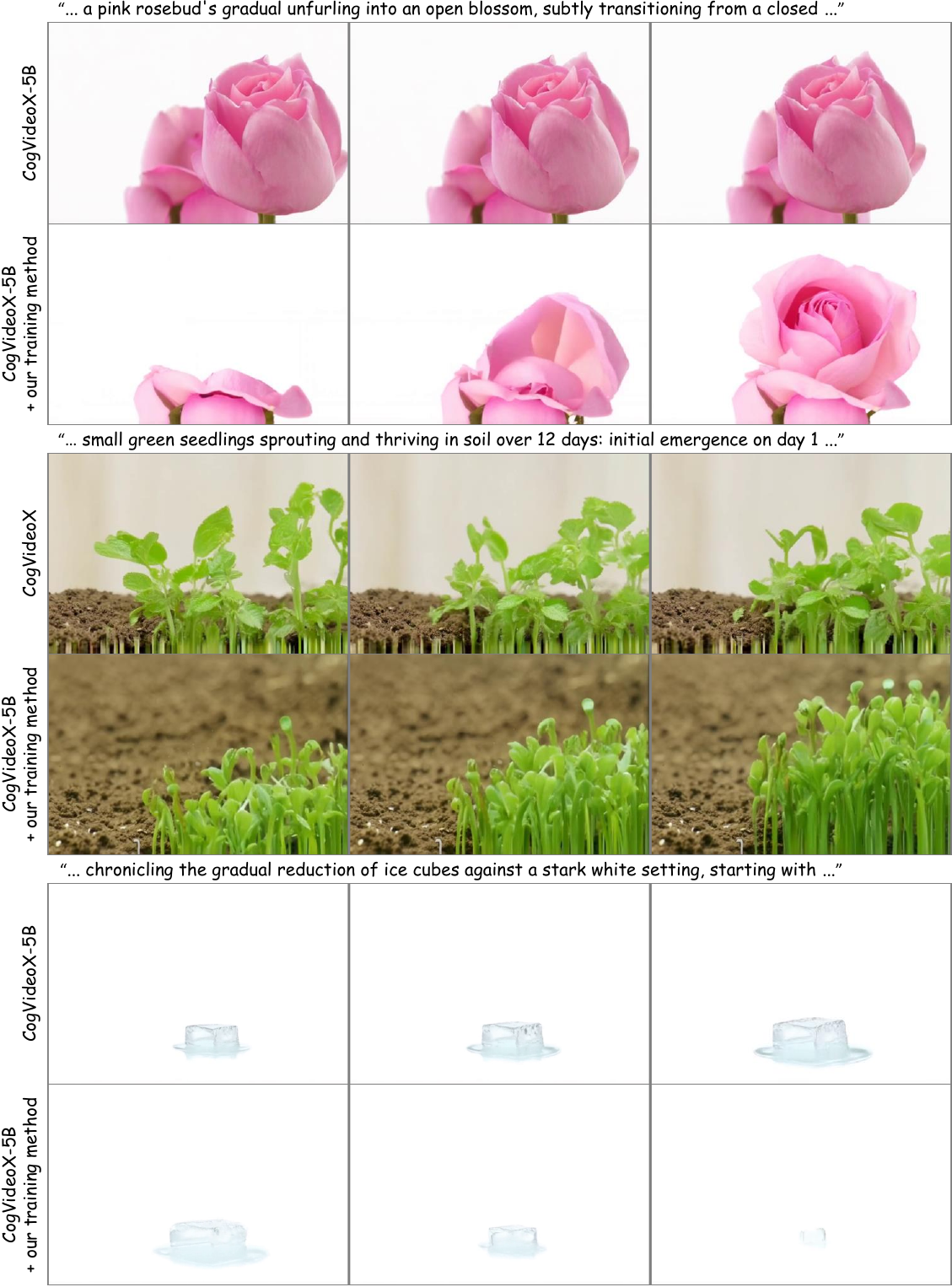}
	\caption{\textbf{Validation of MagicTime Training Recipe on CogVideoX-5B \cite{cogvideox}.} The recipe is effective when scaling to a large video generation model. Baseline \cite{cogvideox} model generates a time-lapse reverse process incorrectly (e.g., ice melting).}
	\label{figure_CogVideoX}
\end{figure}

\jf{
The above baselines are conducted on SD1.5 ($\sim$860M) and Open-Sora-Plan v1.0.0 ($\sim$700M). We further select CogVideoX-5B \cite{cogvideox} to show that the recipe can continue be effective when scaling to larger video generation models. Notably, CogVideoX-5B is among the few recent large models \cite{cogvideox, ltx, mochi} that have publicly released training scripts. To integrate the MagicTime Recipe into the 3D Attention architecture of CogVideoX \cite{cogvideox}, we incorporate MagicAdapter-T and MagicAdapter-S into the first and last halves of the model layers, respectively. According to \cite{sugar}, these two parts primarily correspond to temporal and spatial functions. All other implementation details remain consistent with the official code and this study. The results are shown in Fig. \ref{figure_CogVideoX}. It can be seen that, compared to before fine-tuning, CogVideoX-5B's motion amplitude has significantly increased, and it is capable of generating time-lapse such as seed germination. This validates that the recipe can continue to be effective when scaling to large video generation models. Additionally, even though CogVideoX-5B has 5 billion parameters, it can only promote motion amplitude after fine-tuning for difficult out-of-distribution concepts, similar to MagicTime ($\sim$860M), and cannot generate complete time-lapse.
}

\begin{figure*}[!t]
  \centering
  \includegraphics[width=0.90\linewidth]{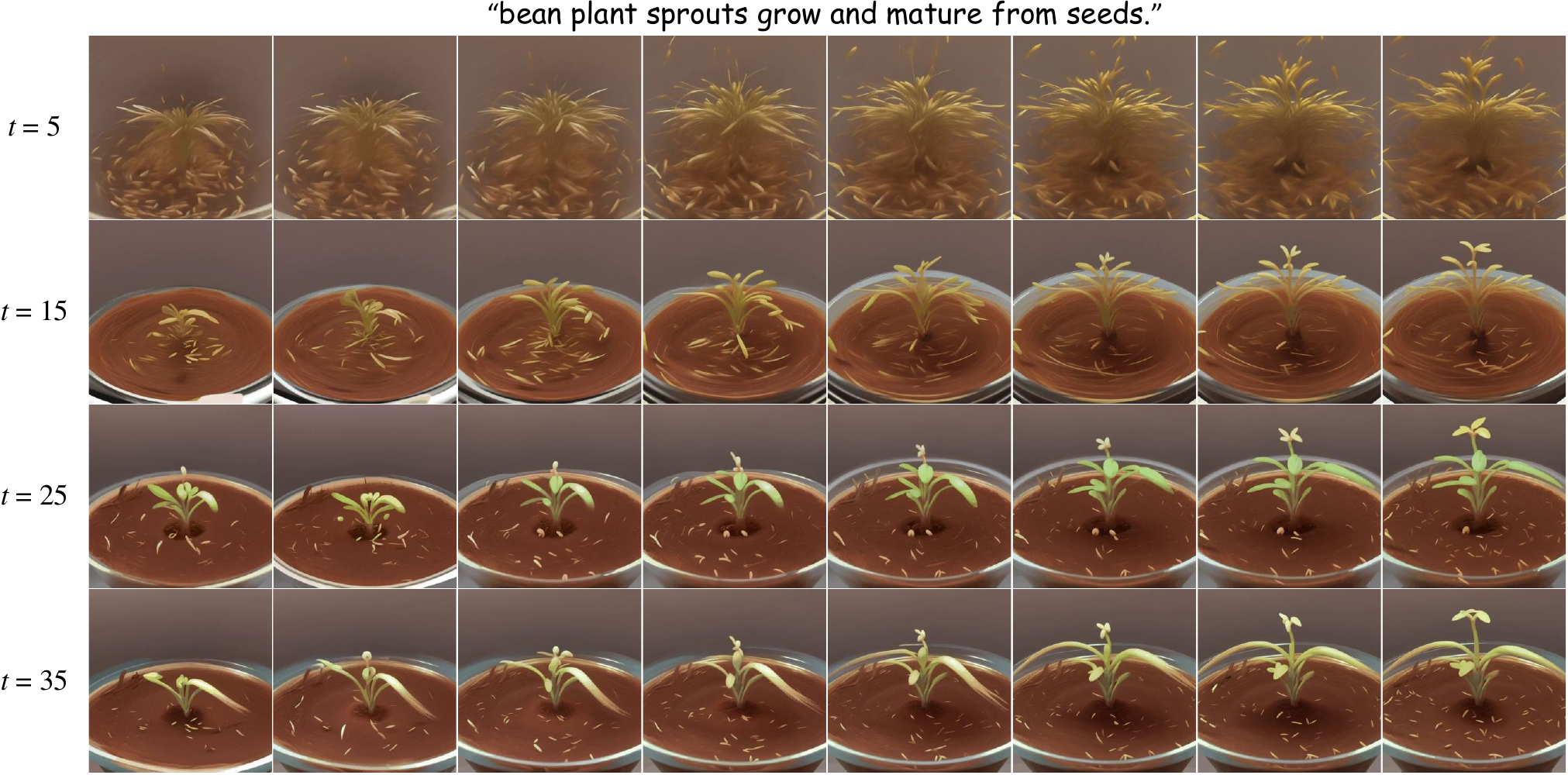}
  \caption{\textbf{Effect of the number of inference steps $t$.} Best results are obtained when $t = 25$.}
  \label{figure_abltion_timesteps}
\end{figure*}

\subsection{Validation of Different Random Seeds}
\begin{figure*}[!t]
    \centering
    \includegraphics[width=0.98\linewidth]{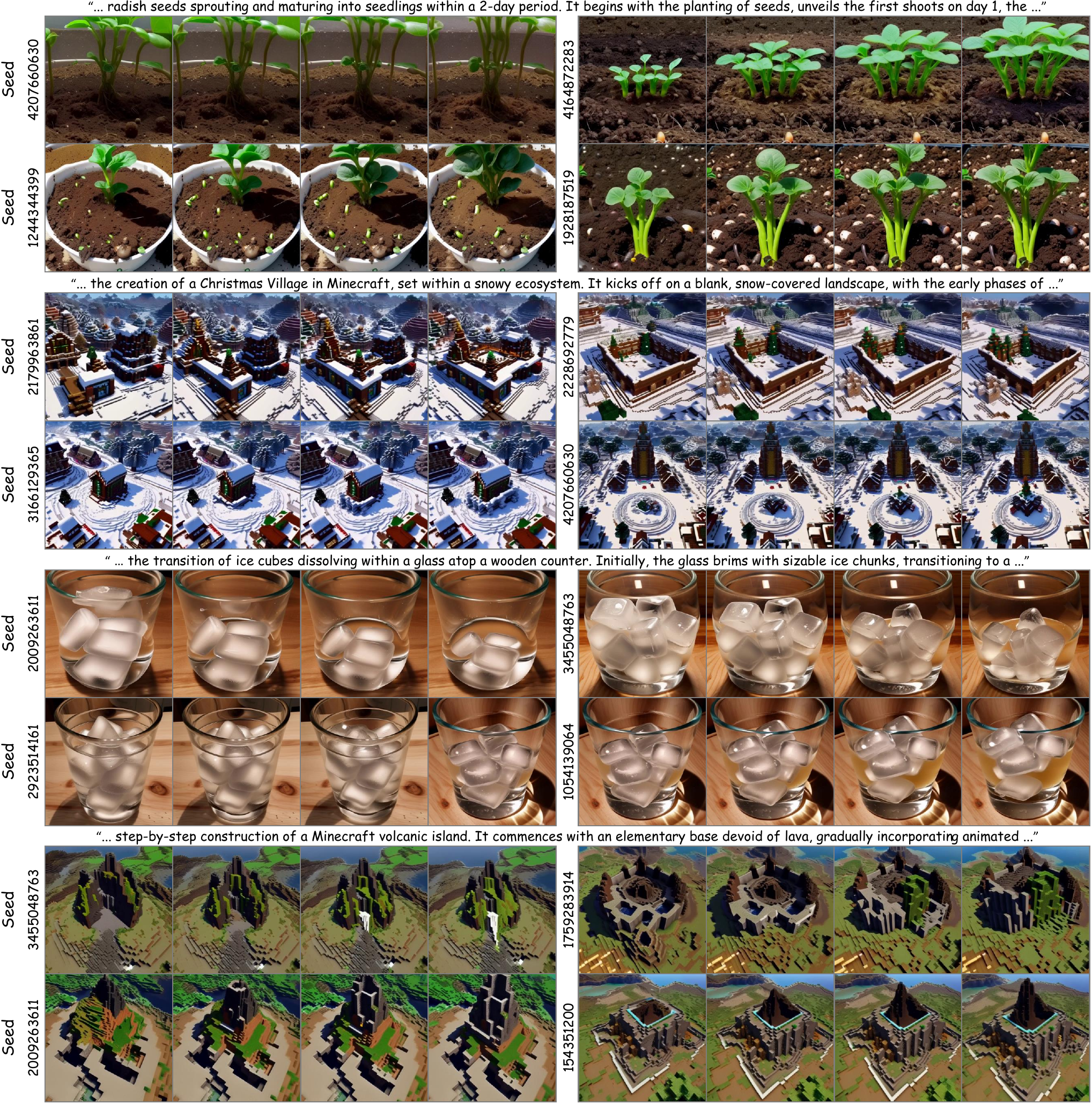}
    \caption{\jf{\textbf{Robustness Test.} MagicTime generates high-quality videos even with different random seeds for the same prompt.}
    }
    \label{figure_different_seed}
\end{figure*}

\jf{
Since the fine-tuning dataset is small, we show some results with different random seeds to test robustness. As shown in Fig. \ref{figure_different_seed}, in Case 1, the plant grows from a seedling to maturity; in Cases 2 and 4, the building components develop from inception to completion; in Case 3, the ice cube gradually melts. These results demonstrate that MagicTime consistently produces high-quality metamorphic videos.
}

\subsection{Validation on Inference Steps}
To assess the impact of the number of inference steps $t$, we conduct an ablation study to investigate how varying $t$ influences the video quality. We set $t = 5, 15, 25, 35$ and apply our approach to the ChronoMagic dataset to obtain metamorphic results. We evaluate Visual Quality (VQ), Frame Consistency (FC), Metamorphic Amplitude (MA), and Text Alignment (TA) for each $t$ value, with examples illustrated in Fig. \ref{figure_abltion_timesteps} using \textit{RcnzCartoon} as a representative case. In terms of VQ, we observe that when $t$ is less than 25, the larger $t$ is, the better VQ becomes. For FC, after $t$ exceeds 5, there is good consistency throughout; for MA, when $t$ equals 25, even if it is increased further, MA only shows minor changes; for TA, regardless of the value of $t$, it can always generate results that are aligned with a text prompt. When $t$ is too small, the diffusion process lacks the necessary flexibility for significant changes, challenging effective generation. Conversely, a larger $t$ value leads to extended reasoning times. In practice, we set $t = 25$ to achieve the best performance in all experiments.

\subsection{Validation on Enhanced Text Prompt}\label{sec: enhanced text prompt}
\begin{figure*}[!t]
  \centering
  \includegraphics[width=0.90\linewidth]{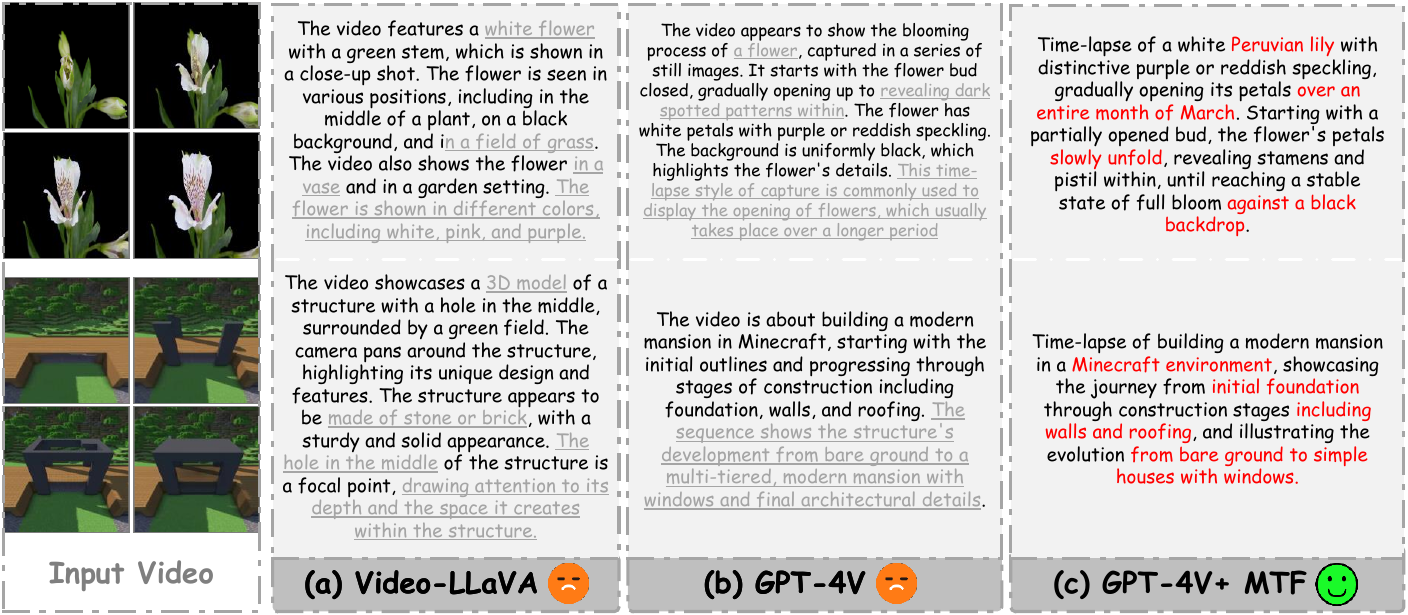}
  \caption{\textbf{Effect of Different Large Multimodal Models and Annotation Strategies for the Generated Captions.} \textcolor{gray}{\textbf{Gray}} indicates inaccuracy or redundancy descriptions, \textcolor{red}{\textbf{red}} indicates precise or detailed descriptions. Text prompts exhibit clear temporal details, abundant object details, and smooth narrative flow when adopting GPT-4V and Multi-view Text Fusion.}
  \label{figure_mtf_ablation}
\end{figure*}

There are various methods exist for video captioning annotation. For example, 360DVD \cite{360DVD} employs BLIP \cite{blip}, which is tailored for captioning tasks, while LanguageBind \cite{Languagebind} uses the multimodal large model OFA \cite{OFA}. However, we observe that both specialized and general models have limitations in comprehending metamorphic videos. We illustrate this by examining the leading multimodal large model Video-LLaVA \cite{Video-llava}, with results in Fig. \ref{figure_mtf_ablation}. It shows Video-LLaVA struggles to capture fine-grained details and often produces irrelevant descriptions that may interfere with prompt instructions. Consequently, we opt for GPT-4V for video annotation. Nevertheless, since GPT-4V only processes keyframe inputs rather than entire videos, it risks losing valuable information. To mitigate this, we introduce Multi-view Text Fusion (MTF) to enhance the quality of video captions. GPT-4V with MTF can generate text prompts with rich temporal details, object specifics, and narrative coherence. We also applied MTF to Video-LLaVA; however, due to its limited ability to follow instructions, the improvements pre- and post-application were relatively minor.

\begin{figure*}[!t]
  \centering
  \includegraphics[width=0.92\linewidth]{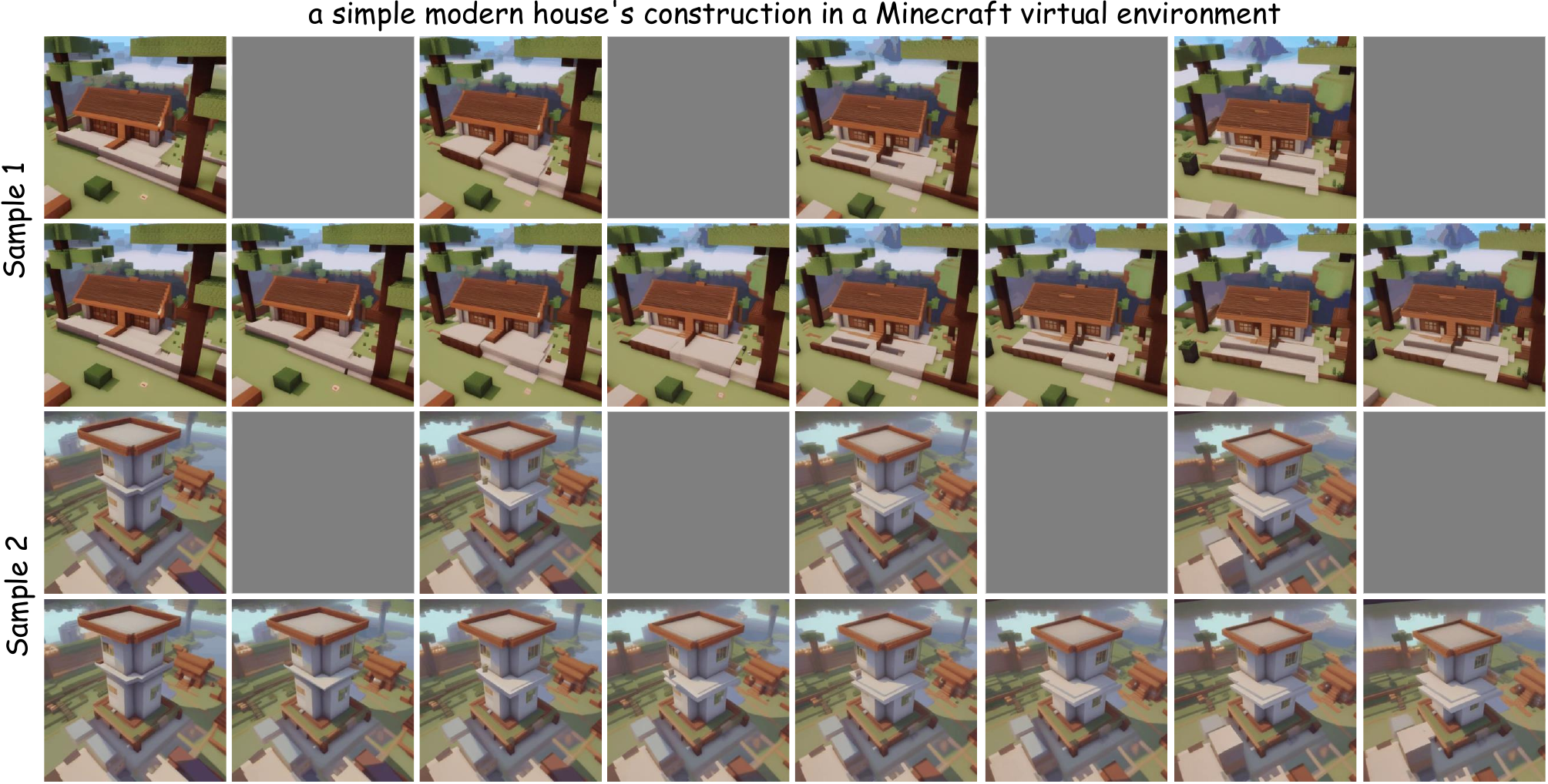}
  \caption{\textbf{Pseudo Long Video Generation.} Our method can be seamlessly converted to a long video generation model by using an interpolation model \cite{amt}.}
  \label{figure_long_video}
\end{figure*}

\subsection{Pseudo Long Video Generation}\label{sec: Pseudo Long Video Generation}
Current text-to-video (T2V) models \cite{lumiere, Latte, animatediff, 360DVD} often face challenges in generating 16-frame videos, which typically feature only camera motion. The generation of longer videos \cite{freenoise} commonly involves multiple forward inferences based on a 16-frame model, followed by concatenating the results. Since our method can produce metamorphic videos within 16 frames, it effectively compresses the content of a longer video into 16 frames. Consequently, combining a frame interpolation model with the outputs of MagicTime can theoretically achieve the effects of longer video generation. We hope this insight will inspire and advance future research in this burgeoning field. We present some examples using the interpolation model \cite{amt}, as illustrated in Fig. \ref{figure_long_video}. It is evident that with MagicTime's capability for metamorphic generation, pseudo-long videos can be produced.

\subsection{Resolution Scalability}\label{sec: Resolution Scalability}
\begin{figure}[!t]
  \centering
  \includegraphics[width=1.0\linewidth]{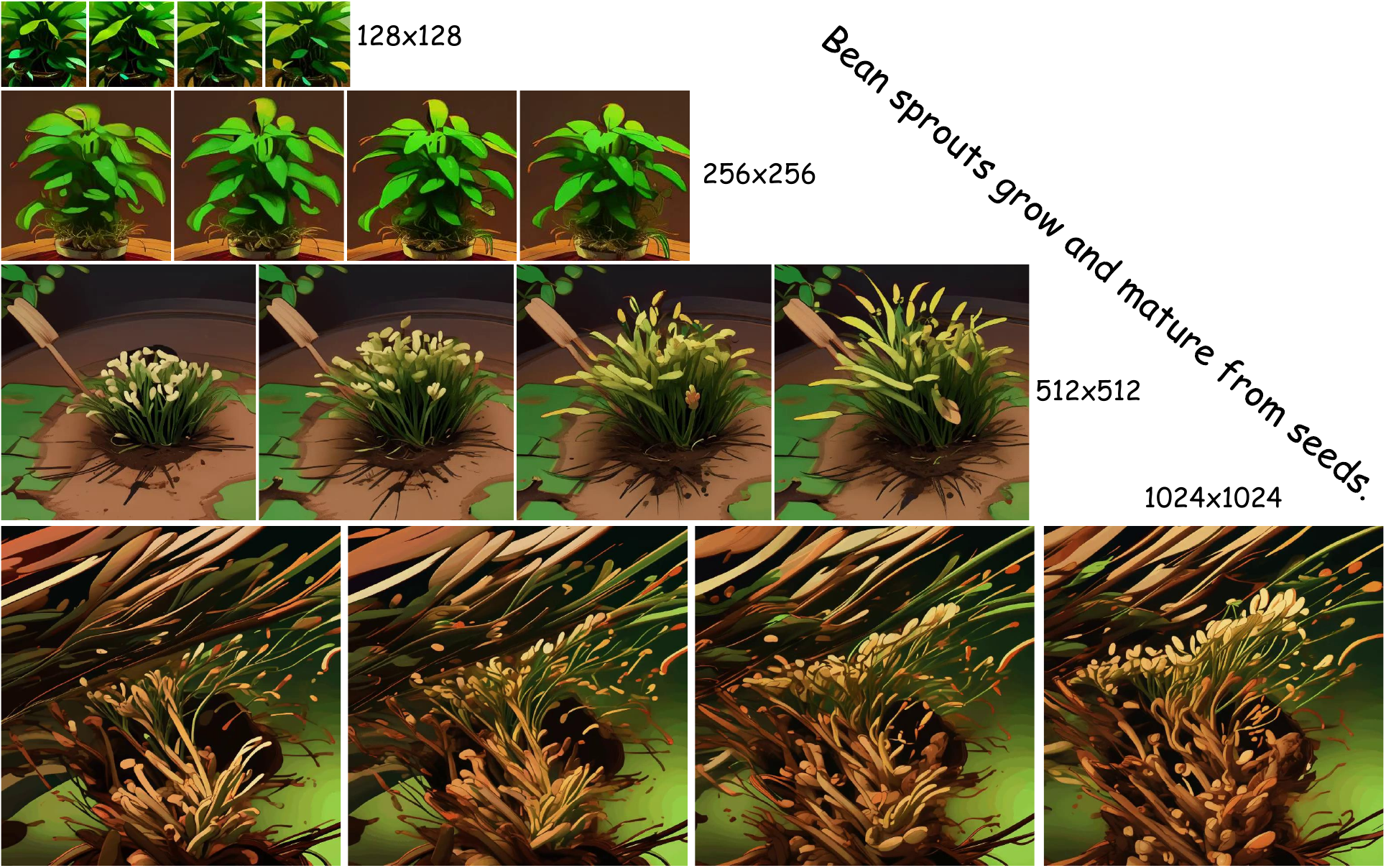}
  \caption{\textbf{Resolution Scalability.} The results demonstrate the generalization capability of our method.}
  \label{figure_scalability}
\end{figure}

Resolution scalability is a significant attribute of the T2V model with a U-Net architecture. To assess whether MagicTime retains this feature, we conducted related experiments, as depicted in Fig. \ref{figure_scalability}. Despite being trained at a resolution of $512\times 512$, our method demonstrates remarkable generalization capabilities across a wide range of resolutions. This scalability is particularly prominent when our model is applied to higher resolutions with $1024 \times 1024$, where it maintains high fidelity and consistency in generated details. Similarly, at lower resolutions, such as $128 \times 128$ and $256 \times 256$, our method preserves structural integrity and aesthetic quality. This robustness may be attributed to the inherent properties of the underlying generative model, Stable Diffusion \cite{LDM}, which we employ in our framework.

\begin{figure*}[!t]
	\centering
	\includegraphics[width=0.91\linewidth]{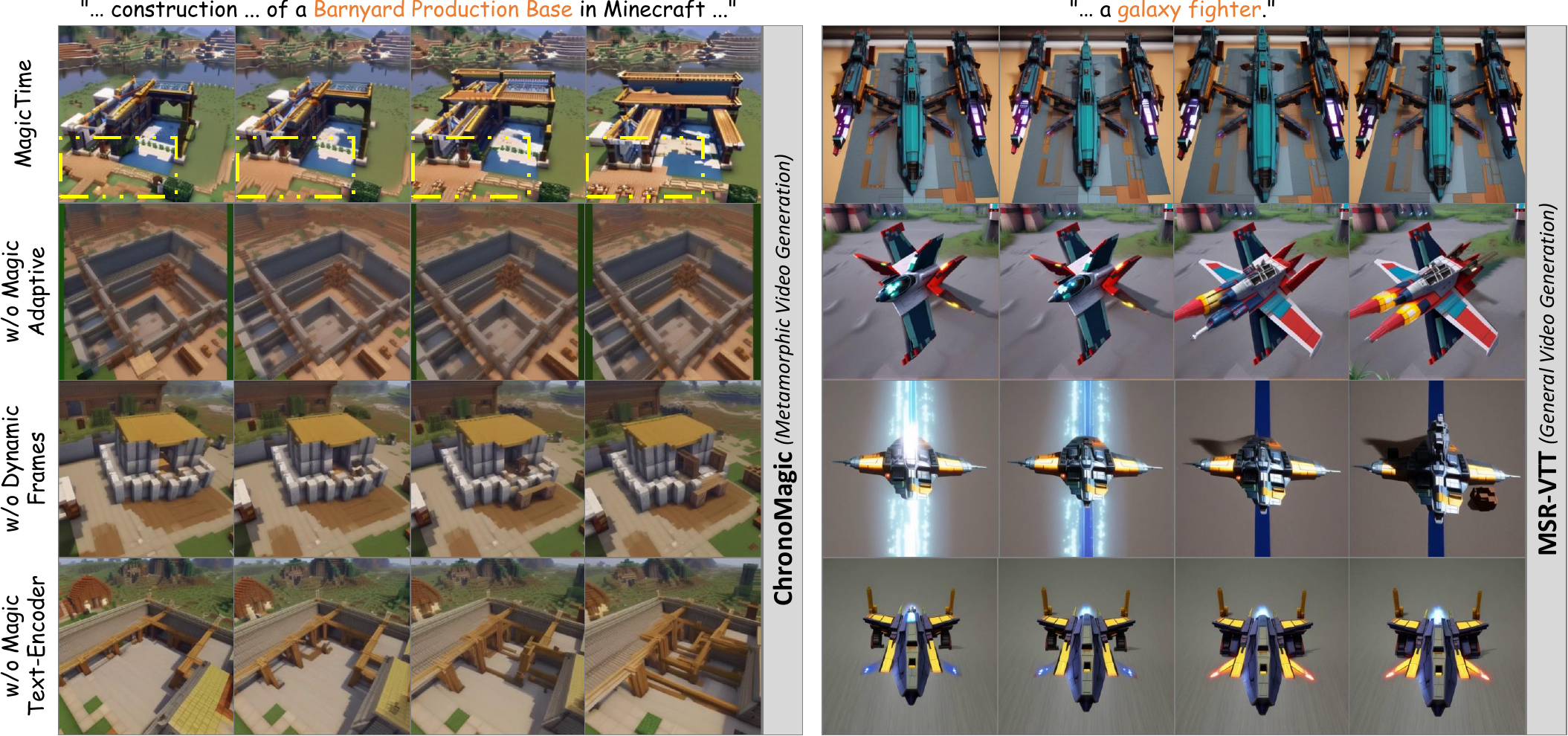}
	\caption{\textbf{More Validation on \textit{Magic Adaptive Strategies}, \textit{Dynamic Frame Extraction} and \textit{Magic Text-Encoder}.} Only by adopting the above three methods can we efficiently encode the metamorphic prior without losing the general video generation capability. The \textcolor{yellow!95!red}{yellow dashed box} corresponds to the content of \textcolor{orange!88!red}{orange word}.}
    \label{figure_ablation_on_general_forgetting_2}
\end{figure*}

\begin{figure}[!t]
  \centering
  \includegraphics[width=0.9\linewidth]{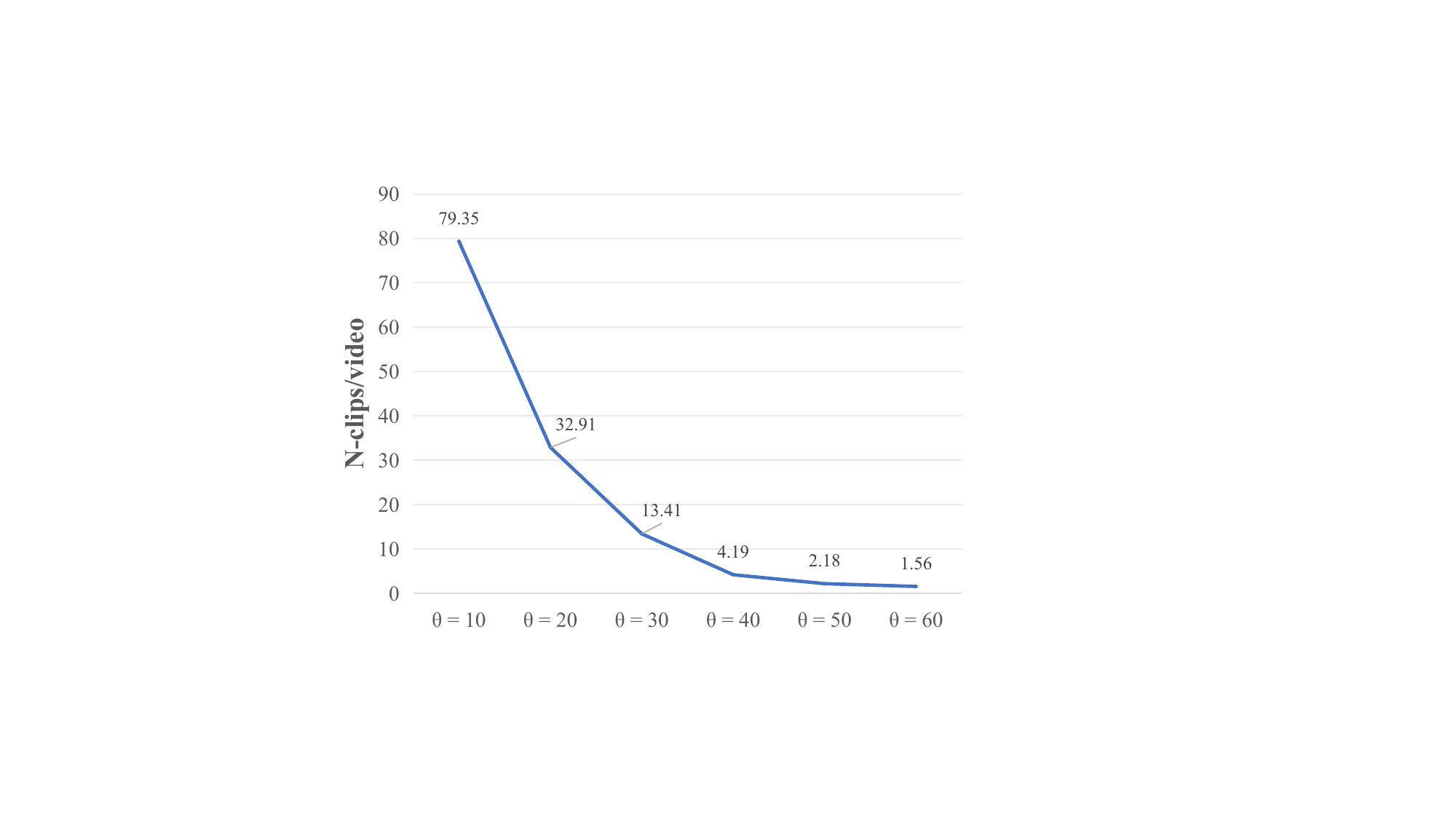}
  \caption{\textbf{Ablation on the settings of $\theta$.} Best results are obtained when $\theta = 40$. N-clips/video represents transitions.}
  \label{figure_hyper_cascade_1}
\end{figure}

\begin{figure*}[!t]
    \centering
    \includegraphics[width=1\linewidth]{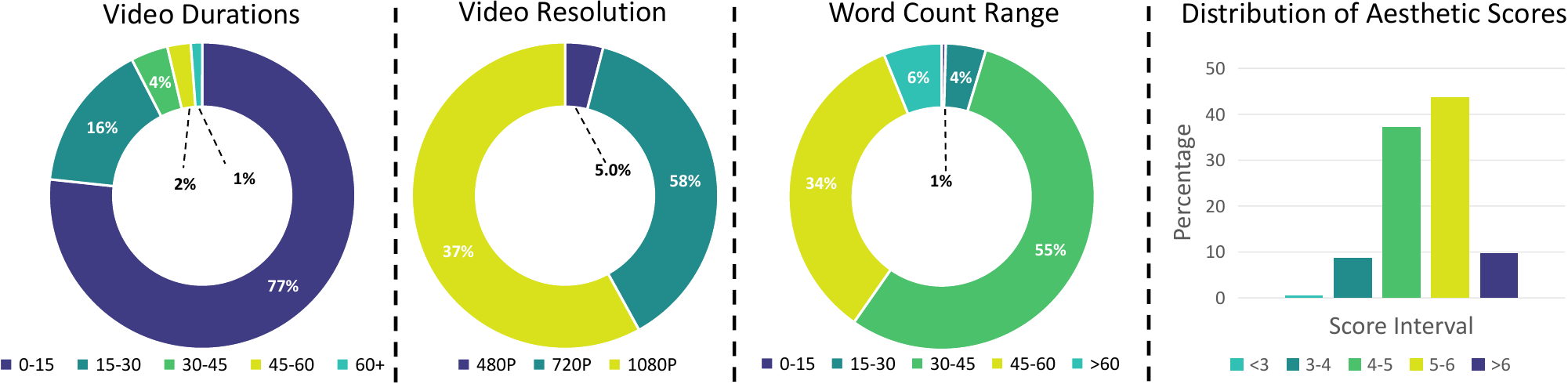}
    \caption{\textbf{Video clips statistics in ChronoMagic dataset.} The dataset includes a diverse range of categories, durations, and caption lengths, with most of the videos being in 1280P resolution.
    }
    \label{figure_dataset_static}
\end{figure*}

\begin{figure*}[!t]
    \centering
    \includegraphics[width=0.80\linewidth]{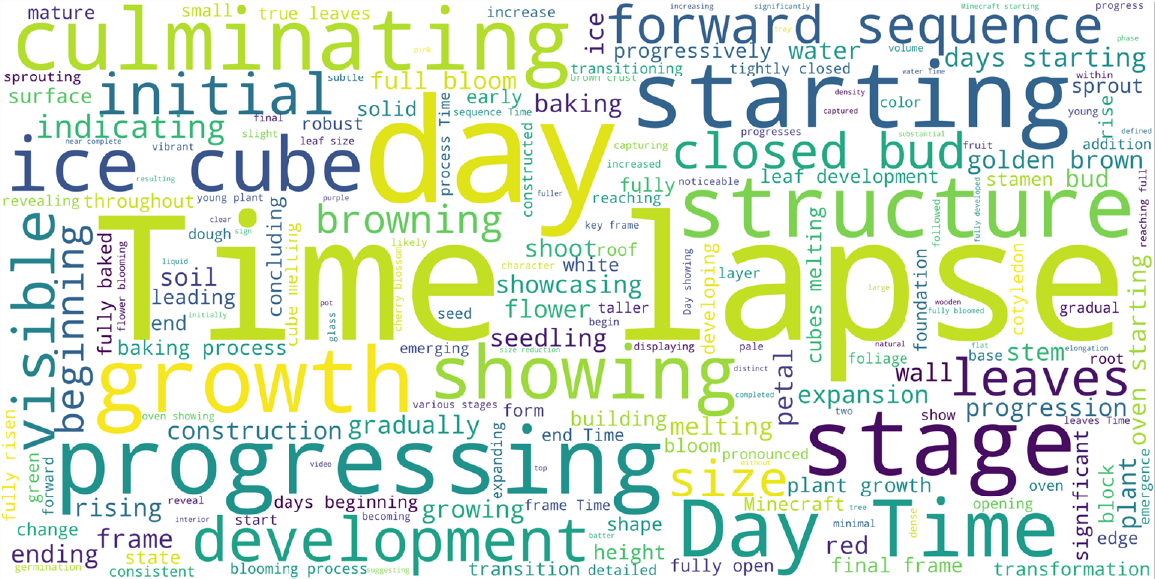}
    \caption{\textbf{The word clouds of the generated captions of the ChronoMagic dataset.} The dataset focuses on changes and processes spanning a large amount of time.
    }
    \label{figure_dataset_word_cloud}
\end{figure*}

\begin{figure*}[!t]
    \centering
    \includegraphics[width=1\linewidth]{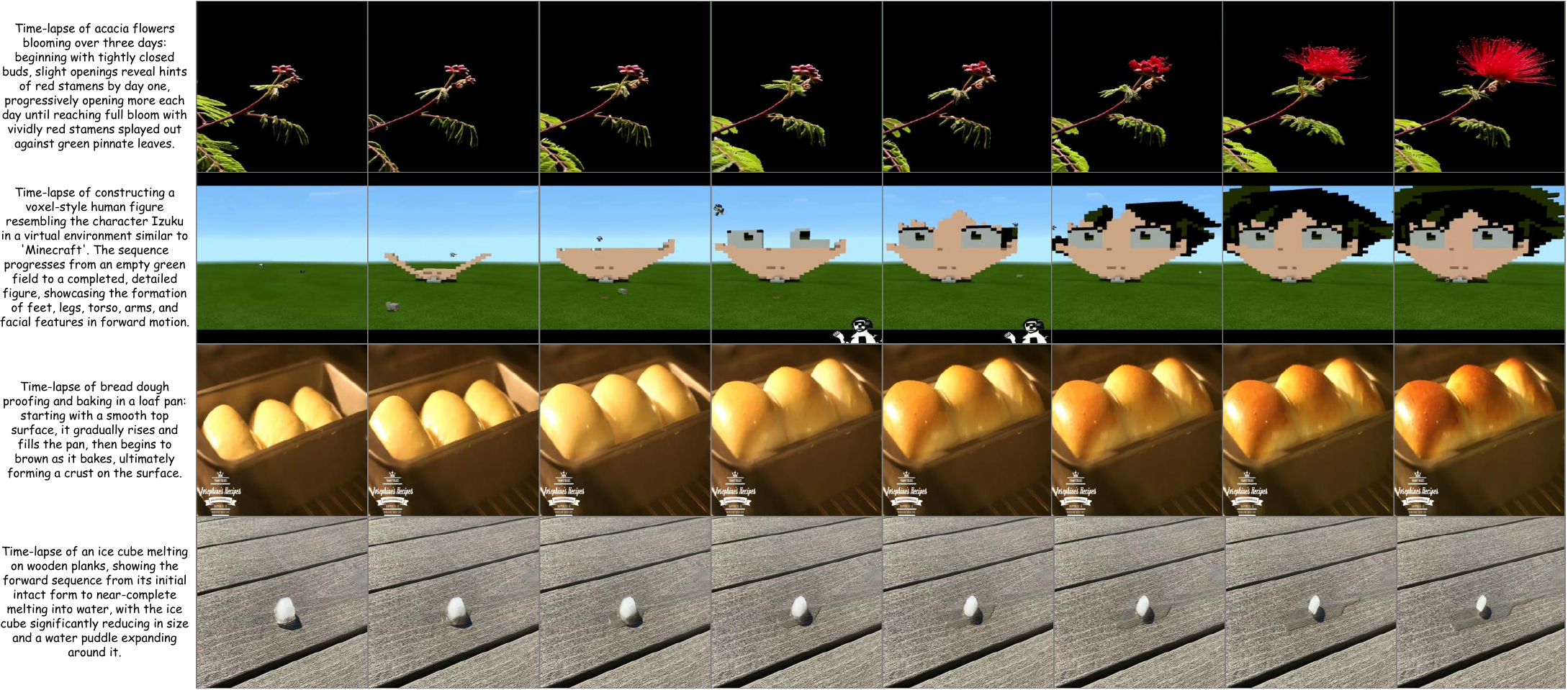}
    \caption{\textbf{Samples from the ChronoMagic dataset.} The dataset consists of metamorphic time-lapse videos, which exhibit physical knowledge, long persistence, and strong variation.
    }
    \label{figure_sample_dataset}
\end{figure*}

\subsection{More Generated Metamorphic Videos}\label{sec: More Cases on Metamorphic Videos}
\begin{figure*}[!t]
    \centering
    \includegraphics[width=1\linewidth]{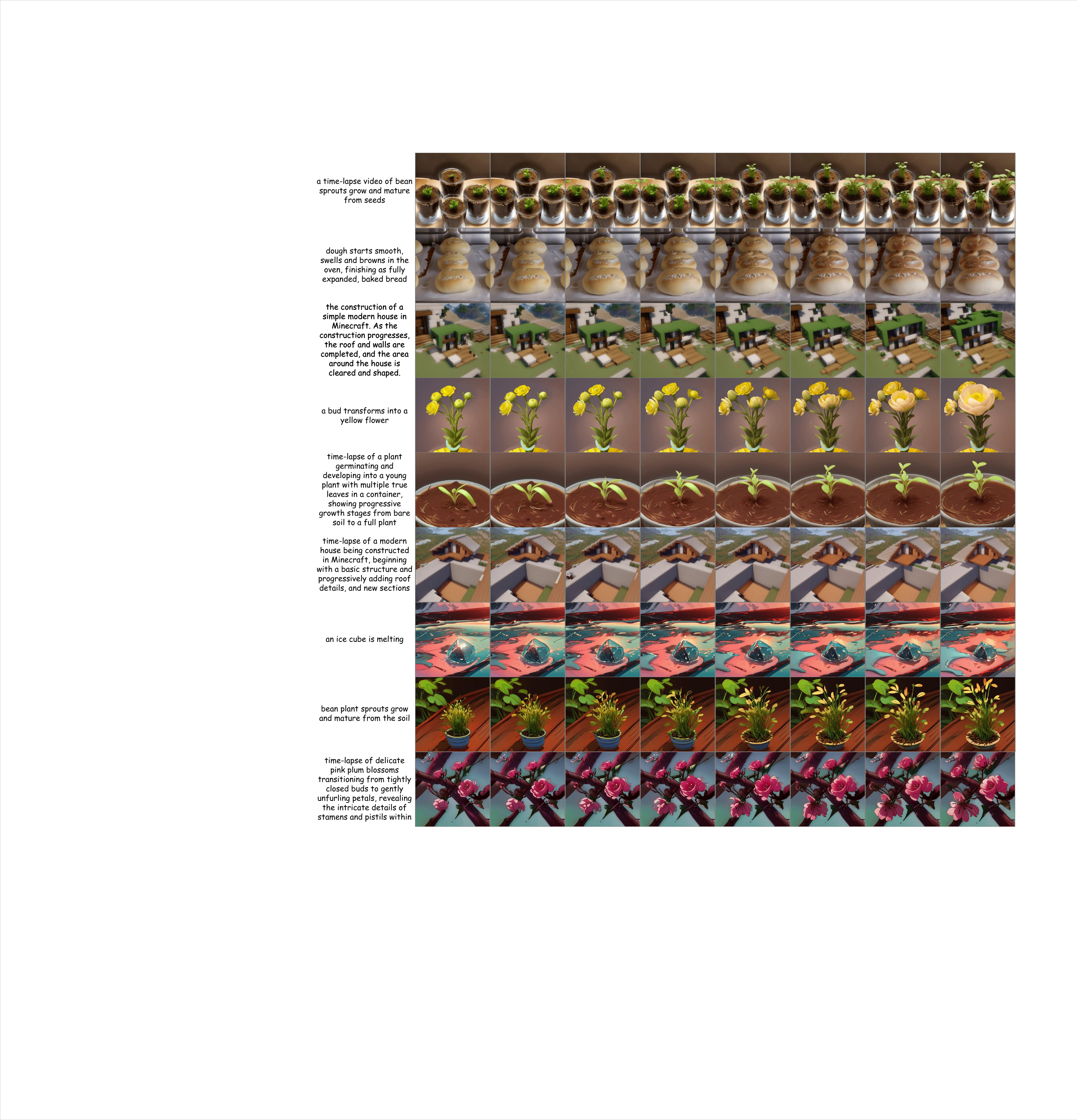}
    \caption{\textbf{Examples of our approach applied to generated metamorphic videos.} Metamorphic videos categories include \textit{RealisticVision} (Row 1-3), \textit{RcnzCartoon} (Row 4-6) and \textit{ToonYou} (Row 7-9). Even with a large state amplitude, the quality, consistency, and semantic relevance of generated videos are excellent.
    }
    \label{figure_main_result_1}
\end{figure*}

\begin{figure*}[!t]
    \centering
    \includegraphics[width=1.0\linewidth]{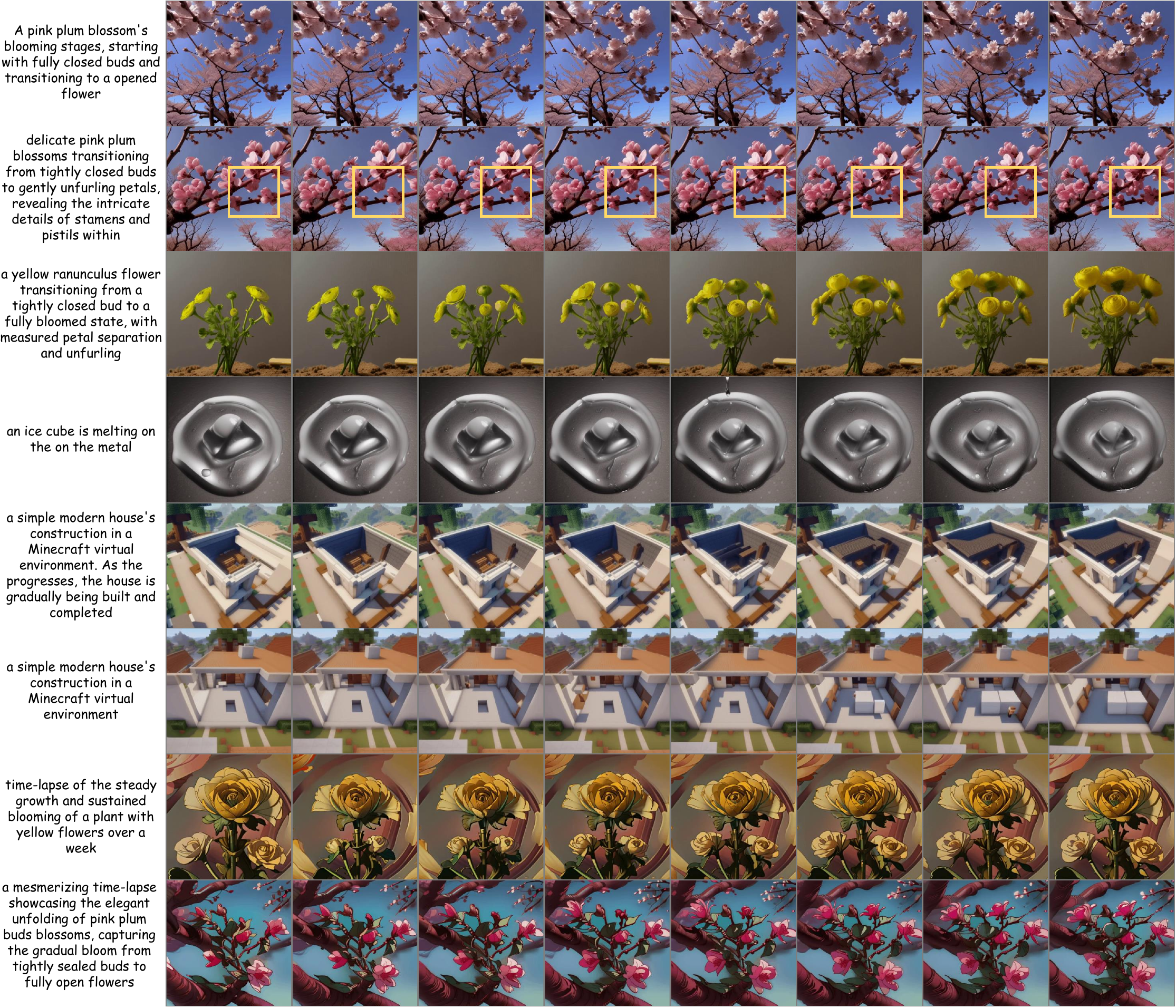}
    \caption{\textbf{More examples of our approach applied to generated metamorphic videos.} Metamorphic videos categories include \textit{RealisticVision} (Row 1-4), \textit{RcnzCartoon} (Row 5-6) and \textit{ToonYou} (Row 7-8). Even with a large state amplitude, the quality, consistency, and semantic relevance of generated videos are excellent.
    }
    \label{figure_main_result_2}
\end{figure*}

\begin{figure*}[!t]
    \centering
    \includegraphics[width=1\linewidth]{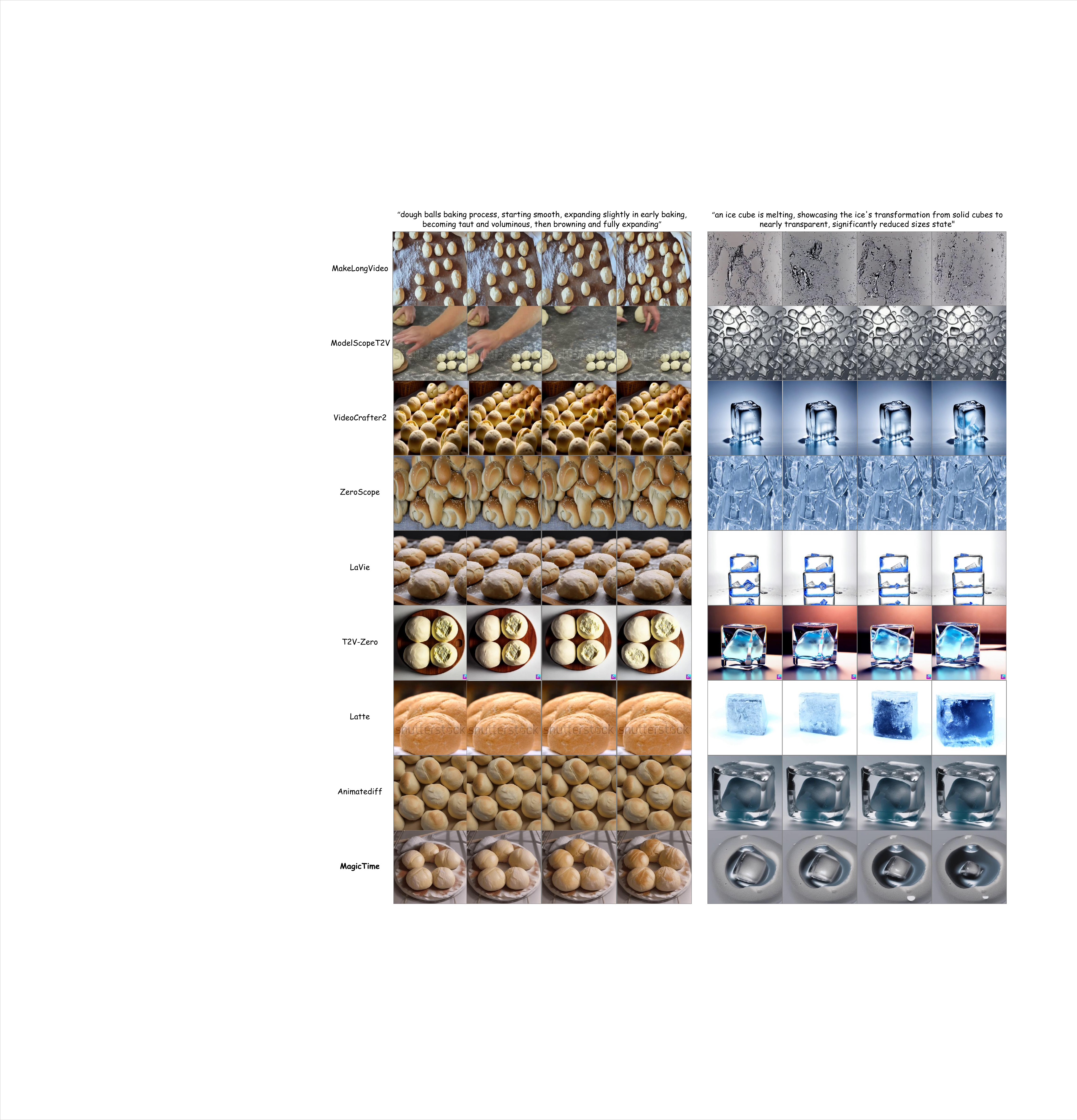}
    \caption{\textbf{More qualitative comparison of metamorphic video generation.} MagicTime achieves better results compared to the current leading T2V models.
    }
    \label{figure_comparision_CTL_2}
\end{figure*}

Due to space constraints, to demonstrate the robustness and generality of our approach, we present several text-guided metamorphic video generation results in Fig. \ref{figure_main_result_1} and Fig. \ref{figure_main_result_2}. We also present additional metamorphic videos generated by MagicTime and other baseline methods in Fig. \ref{figure_comparision_CTL_2}. It is clear that MagicTime consistently surpasses existing state-of-the-art Time-to-Video (T2V) models across various scenarios, from melting ice to the expansion of bread. These results showcase a robust understanding of physical phenomena encoded within the model. Even when the textual prompt requires the generation of content spanning long periods, the generated videos maintain high quality, consistency, and semantic relevance. 


\subsection{More Validation on MagicTime Recipe}\label{sec: More Validation on Training Scheme and Catastrophic Forgetting}
We provide more cases to illustrate the gains of the proposed MagicTime Recipe, as shown in Figure \ref{figure_ablation_on_general_forgetting_2}. Specifically, the model trained without the \textit{Magic Adaptive Strategy} exhibits significant flickering (e.g., in Minecraft and fighter scenes). We hypothesize that this issue arises from the substantial differences between the metamorphic process and conventional video generation. Additionally, the absence of \textit{Dynamic Frame Extraction}, which employs uniform frame extraction instead, compromises the model's overall video generation capability. This suggests that the model learns the metamorphic process by fitting to data rather than encoding physical information, leading to erratic changes in the generated general videos (e.g., continuously expanding spaceship shapes). The \textit{Magic Text Encoder} showcases an enhanced understanding of prompts; for instance, only MagicTime accurately generates the content of the yellow box as specified by the orange text, while other models only grasp coarse-grained information.

\begin{figure*}[!t]
\centering
    \includegraphics[width=0.85\linewidth]{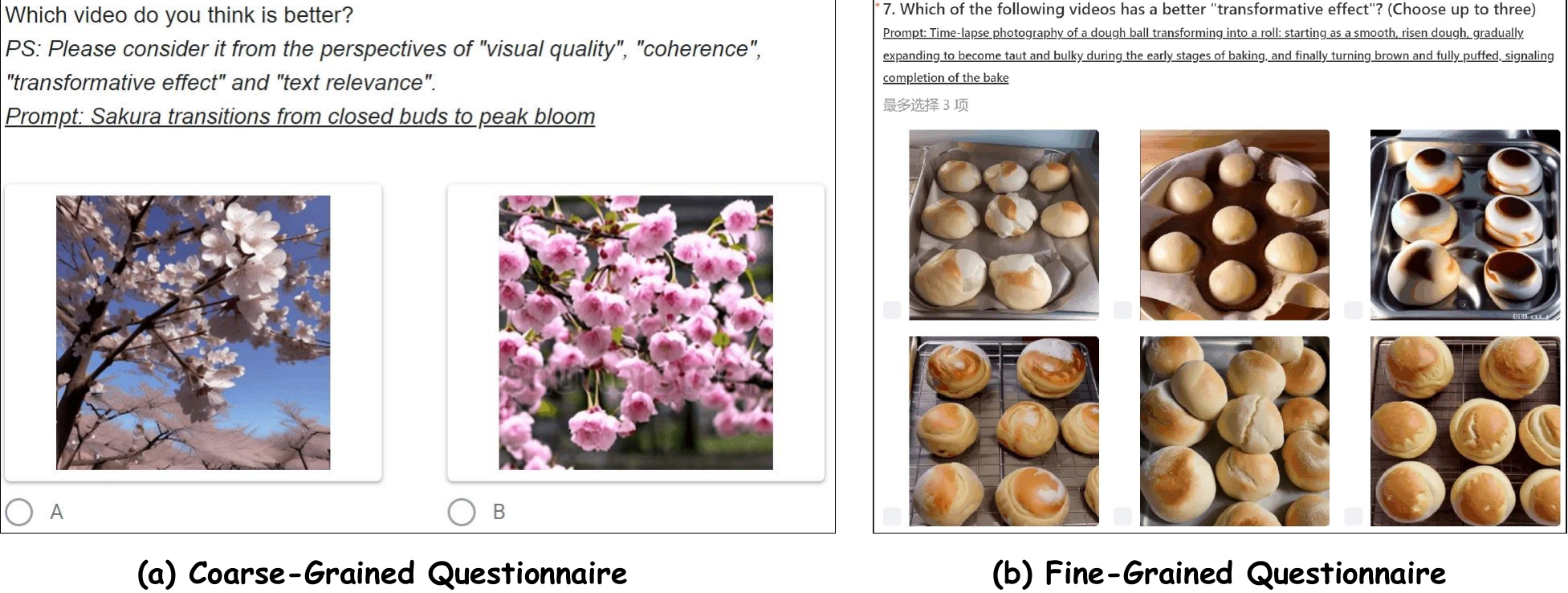}
    \caption{\textbf{Visualization of the Questionnaire for Human Evaluation.}}
\label{figure_questionaire}
\end{figure*}

\begin{figure}[!t]
    \centering
    \includegraphics[width=1\linewidth]{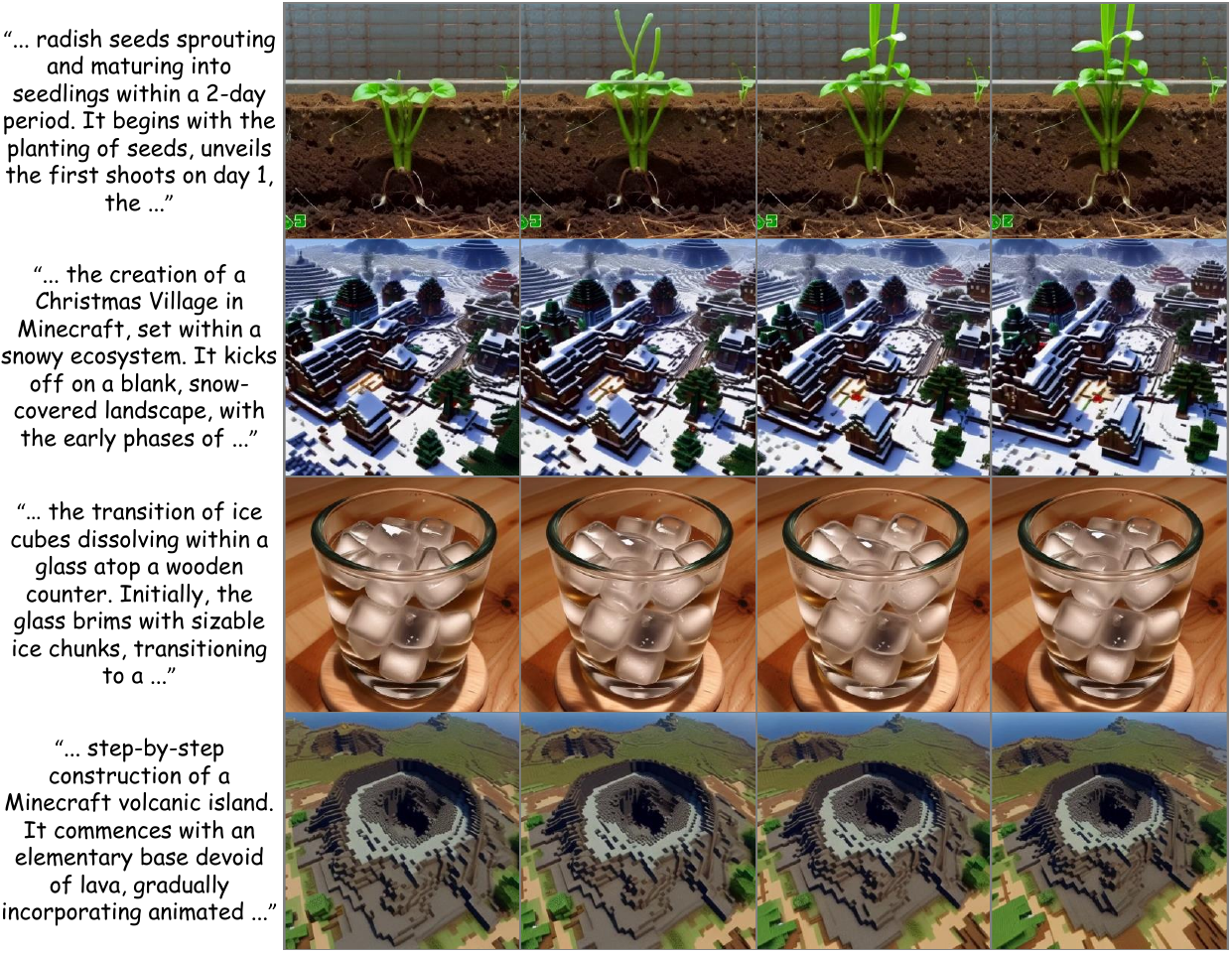}
    \caption{\jf{\textbf{Failure Cases.} MagicTime may generate videos with watermarks or merely camera movements.}}
    \label{figure_failure_cases}
\end{figure}

\begin{figure}[!t]
  \centering
  \includegraphics[width=0.55\linewidth]{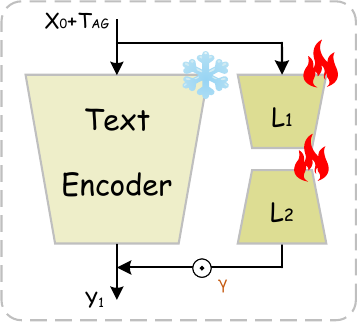}
  \caption{\textbf{Structure of the Magic Text-Encoder}. \(X_0\) represents the original text prompt, while \(T_{AG}\) is the additional metamorphic tags. \(L_1\) and \(L_2\) are trainable parameters.}
  \label{figure_magicTextEncoder}
\end{figure}

\subsection{More Hyperparameter Analysis}\label{sec: More Hyperparameter Analysis}
In the Cascade Preprocessing module, we detect video transitions based on the grayscale difference between two frames. The choice of the difference threshold $\theta$ significantly impacts the ability to identify transitions. Therefore, we conducted ablation experiments with varying $\theta$ values to find optimal values, as depicted in Fig. \ref{figure_hyper_cascade_1}. The actual average number of transitions is approximately six. When $\theta$ is too large, many frames are erroneously identified as transitions. Conversely, when $\theta$ is too small, it results in missed detections. It can be seen that optimal results are achieved when $\theta$ is set to 40.

\section{Additional Visualization}\label{sec: Additional Visualization}
\subsection{Structure of Magic Text-Encoder}\label{sec: Structure of Magic Text-Encoder}
The structure of the Magic Text-Encoder, depicted in Fig. \ref{figure_magicTextEncoder}, comprises the original CLIP Text Encoder and several newly added convolutional layers \(L_1\) and \(L_2\). During the training process, only the newly added modules (e.g., \(L_1\) and \(L_2\)) are trained; subsequently, the results are weighted and combined with the output of the CLIP Text Encoder \(X_0\).

\subsection{Samples from the ChronoMagic Dataset}\label{sec: Samples from the ChronoMagic}
In this work, we compiled a collection of time-lapse videos from the Internet to create a metamorphic video-text dataset containing 2,265 videos, named the ChronoMagic. Fig. \ref{figure_sample_dataset}, Fig. \ref{figure_dataset_static} and Fig. \ref{figure_dataset_word_cloud} showcases statistics and samples from the dataset, which primarily consists of metamorphic time-lapse videos of plants, buildings, ice, food, and other categories. We plan to scale up the dataset to include additional categories and a larger number of videos in the future.

\subsection{Questionnaire Visualization}\label{sec: Questionnaire Visualization}
To enhance the reliability of the experiment, we developed both coarse-grained and fine-grained questionnaires during the data collection phase and invited 200 participants to provide their responses. The coarse-grained questionnaire required participants to assess these four criteria and select the video that best met their requirements, as shown in Fig. \ref{figure_questionaire}a. As this questionnaire was more user-friendly, a total of 300 questionnaires were collected for all surveys related to this work. The fine-grained questionnaire asked participants to evaluate each video set's visual quality, frame consistency, metamorphic amplitude, and text alignment, as shown in Fig. \ref{figure_questionaire}b. Due to the time-consuming nature of this questionnaire, the final number of responses collected was insufficient, so it is not used in the main text. 

\section{Additional Statement}\label{sec: Additional Statement}
\subsection{Limitations}
\jf{First, its performance depends on the foundational models of general video generation, as these parameters are frozen during training. While this ensures generative capabilities are preserved, it also limits the potential of our method. It may produce videos with watermarks or simply camera movements for in-distribution categories, as shown in Fig. \ref{figure_failure_cases}. Second, although we have verified that time-lapse videos can encode more physical priors into the model, the limited scale of data prevents MagicTime from generating metamorphic videos of all categories, as shown in Fig. \ref{figure_Out-of-Distribution Concepts}.}



\subsection{Ethics Statement}
MagicTime can generate not only metamorphic videos with large time spans and high levels of realism but also high-quality general videos. Potential negative impacts exist as the technique can be used to generate fake video content for fraud. We note that any technology can be misused.

\subsection{Reproducibility Statement} First, we have explained the implementation of each component of MagicTime in detail in Section~\ref{method}. Second, we have clarified training and inference details in Section~\ref{sec:exp setting}. Finally, all datasets and codes used in this work will be open-source.


\end{document}